\documentclass{article}
\PassOptionsToPackage{numbers, compress}{natbib}
\usepackage[preprint]{neurips_2020}
\usepackage[utf8]{inputenc}
\usepackage[T1]{fontenc}
\usepackage{hyperref}
\usepackage{url}
\usepackage{booktabs}
\usepackage{amsfonts}
\usepackage{nicefrac}
\usepackage{microtype}
\usepackage{amsmath}
\usepackage{amsthm}
\usepackage{amssymb}
\usepackage{amsfonts}
\usepackage{graphicx}
\usepackage{subfigure} 
\usepackage{stmaryrd}
\usepackage[capitalise]{cleveref}
\usepackage{capt-of}
\usepackage[misc]{ifsym}

\hypersetup{
    colorlinks=true,
    linkcolor=blue,
    filecolor=magenta,      
    urlcolor=cyan,
}

\usepackage{xr-hyper}
\usepackage{hyperref}

\newtheorem{theorem}{Theorem}

\newcommand{\ie}{\textit{i}.\textit{e}., }
\newcommand{\eg}{\textit{e}.\textit{g}., }

\title{Target Consistency for Domain Adaptation:\\ when Robustness meets Transferability}

\author{
  Yassine Ouali$^*$$^\dag$(\Letter), Victor Bouvier\thanks{Equal contribution} $\ ^\dag$$^\ddagger$(\Letter), Myriam Tami$^\dag$, and Céline Hudelot$^\dag$\\
  $^\dag$Université Paris-Saclay, CentraleSupélec, Mathématiques et Informatique pour la \\ Complexité et les Systèmes, 91190, Gif-sur-Yvette, France, \\ {\footnotesize{\texttt{\{yassine.ouali, myriam.tami, celine.hudelot\}@centralesupelec.fr}}}\\
  $^\ddagger$Sidetrade, 114 Rue Gallieni, 92100, Boulogne-Billancourt, France\\
  {\footnotesize{\texttt{vbouvier@sidetrade.com}}}
}

\begin{document}

\maketitle

\begin{abstract}

Learning \textit{Invariant Representations} has been successfully applied for reconciling a source and a target domain for Unsupervised Domain Adaptation. %In this paper, we investigate the robustness of such methods under the prism of the cluster assumption. Our findings show that the cluster assumption is violated in the target domain despite being maintained in the source domain. 
By investigating the robustness of such methods under the prism of the cluster assumption, we bring new evidence that invariance with a low source risk does not guarantee a well-performing target classifier. More precisely, we show that the cluster assumption is violated in the target domain despite being maintained in the source domain, indicating a lack of robustness of the target classifier. To address this problem, we demonstrate the importance of enforcing the cluster assumption in the target domain, named \textit{Target Consistency} (TC), especially when paired with \textit{Class-Level InVariance} (CLIV). Our new approach results in a significant improvement, on both image classification and segmentation benchmarks, over state-of-the-art methods based on invariant representations. Importantly, our method is flexible and easy to implement, making it a complementary technique to existing approaches for improving transferability of representations.

\end{abstract}

\section{Introduction}
Deep learning (DL) models often show a weak ability to generalize on samples significantly different from those seen during training \cite{beery2018recognition, arjovsky2019invariant, geva2019we}. This inability to generalize out of the training distribution presents a significant obstacle to a controlled and safe deployment of DL models in real-world systems \cite{amodei2016concrete, marcus2020next}. To bridge the distribution gap, Unsupervised Domain Adaptation (UDA) \cite{amodei2016concrete,pan2009survey} leverages labeled samples from a well-known domain, referred to as \textit{source}, to generalize on a \textit{target} domain, where only unlabeled samples are available. If labelling functions are equal across domains, a situation known as the \textit{covariate shift}, adaptation can be performed by weighting sample contributions in the loss  \cite{sugiyama2007covariate, sugiyama2008direct, gretton2012kernel, quionero2009dataset}. However, for high dimensional data, such as text or image, it is unlikely that source and target distributions share enough statistical support to compute weights \cite{johansson2019support}. Learning domain \textit{Invariant Representations} \ie representations for which it is impossible to distinguish the domain they were sampled from, can bring together two domains which are different in the input space \cite{ganin2015unsupervised, long2015learning}. This fundamental idea, and the corresponding theoretical target risk \cite{ben2007analysis, ben2010theory}, has led to a wide variety of methods for adapting deep classifiers to new domains \cite{long2016unsupervised, long2017deep, long2018conditional}.

Nevertheless, the invariance of representations does not always guarantee a low target risk. For instance, in the case of images, aligning source and target backgrounds reduces domain discrepancy of representations but is unlikely to improve the model in the target domain. Even worse, it may incorrectly align source and target classes if the background is incorrectly correlated with a given class due to some collection bias \cite{beery2018recognition, arjovsky2019invariant}, phenomenon known as \textit{negative transfer} \cite{torrey2010transfer}. %We refer to this phenomenon as  \textit{negative transfer} \cite{torrey2010transfer}, where knowledge from the source domain is transferred by hook or by crook and worsens target generalization. 
Some theoretical works have investigated the question of negative transfer when label shift between source and target domains is observed \cite{zhao2019learning, johansson2019support}, revealing a fundamental trade-off between invariance and ability of predictions to match the true target label distribution \cite{zhao2019learning}. Prior works address this trade-off by relaxing domain invariance with weighted representations \cite{cao2018partial, you2019universal, long2018conditional, combes2020domain}. However, learning invariant, but transferable representations, remains an open problem. One of the main hurdles is the negative impact invariance has on discriminability, resulting in sub-optimal and sensitive target classification.
%However, the difficulty of learning transferable representations goes beyond the label shift situation. 
Recently,  significant signs of progress have been achieved; (\textit{i}) by exposing to a domain discriminator the multi-linear conditioning between representations and predictions \cite{long2018conditional}, (\textit{ii}) by penalizing high singular values in a batch of representations to enhance discriminability \cite{chen2019transferability} or (\textit{iii}) by generating intermediate transferable representations for reducing domain discrepancy \cite{liu2019transferable}. 

In the present work, we aim to provide a new understanding of the transferability of representations through the prism of the cluster assumption, a well-known semi-supervised learning paradigm. The cluster assumption states that if samples are in the same cluster in the input space, they are likely to be of the same class. When enforced on unlabeled samples, the model benefits from a significant gain in generalization \cite{4787647,sohn2020fixmatch,xie2019unsupervised} and robustness \cite{carmon2019unlabeled, hendrycks2019augmix}. We show that enforcing the cluster assumption in the target domain, named \textit{Target Consistency} (TC), with domain invariant representations goes beyond the role of a regularizer for high capacity features extractor as described in \cite{shu2018dirt}. %When enforcing the cluster assumption in the target domain, referred to as \textit{Target Consistency} (TC), we demonstrate that it improves representations discriminability by pushing them far from the decision boundary. 
%Crucially, we reveal that the gain induced by target consistency interacts substantially with class-level domain invariant representations obtained by fooling one discriminator per class. 
Crucially, we reveal that class-level invariance maximizes the gains induced by Target Consistency. By fooling one discriminator per predicted class, we promote positive interaction between TC and Class-Level InVariance (CLIV).
Our contributions are the following: (\textit{i}) Through an in-depth empirical analysis, we show that domain invariance induces a significant model sensitivity to perturbations in the target domain. It indicates that invariance is achieved by disregarding principles of robustness. Such evidence motivates our interest in enforcing the cluster assumption for improving the transferability of domain invariant representations. (\textit{ii})  
To amplify the effect of TC, we perform class-level invariance (CLIV) while enforcing the cluster assumption, promoting positive feedback between decision boundary updates and representation alignment. (\textit{iii}) We show with extensive experiments on both classification and segmentation datasets that we reach state-of-the-art performances for methods based on invariant representations.

\section{On the Vulnerability of Invariant Representations}

\paragraph{Preliminaries.} \textit{Domain Adaptation} (DA) introduces two domains, the \textit{source} and the \textit{target} domains, on the product space $\mathcal X \times \mathcal Y$ where $\mathcal X$ is the input space and $\mathcal Y$ is the label space. Those domains are defined by their specific joint distributions of inputs $\mathbf x \in \mathcal X$ and labels $\mathbf y \in \mathcal Y$, noted $p(\mathbf x^s, \mathbf y^s)$ and $q(\mathbf x^t,\mathbf y^t)$, respectively. We refer to quantities involving the source and the target as $s$ and $t$, respectively, with exponent notation. Considering a hypothesis class $\mathcal H$, subset of functions from $\mathcal X$ to $\mathcal Y$, DA aims to learn $h \in\mathcal H$ which performs well in the target domain \textit{i.e.} has a small target risk $\varepsilon^t(h) := \mathbb E_{(\mathbf x^t, \mathbf y^t) \sim p}[\ell(h(\mathbf x^t), \mathbf y^t)]$ where $\ell$ is a given loss. \textit{Unsupervised Domain Adaptation} (UDA) considers the case where labeled samples are available in the source domain while the target domain is only represented with unlabeled samples. Learning \textit{Domain Invariant Representations} is a key idea for reconciling two non-overlapping data distributions \cite{ganin2015unsupervised, long2015learning}. A mainstream approach consists in learning a representation with a deep feature extractor such that a domain discriminator can not distinguish target from source samples \cite{ganin2015unsupervised}. Provided a representation class $\Phi$, a subset of functions from the input space $\mathcal X$ to a representation space $\mathcal Z$, a classifier class $\mathcal G$, a subset of functions from $\mathcal Z$ to $\mathcal Y$, and noting $\mathcal H := \mathcal G \circ \Phi := \{g \circ \varphi; g \in \mathcal G, \varphi \in\Phi\}$, representations $\varphi \in \Phi$ are learned by achieving a trade-off between minimizing source classification error and fooling a domain discriminator \cite{ganin2015unsupervised}, expressed as a function from $\mathcal Z$ to $[0,1]$. The role of representations in UDA has been theoretically investigated by Ben-David \textit{et al.} in \cite{ben2007analysis}, and extended in \cite{ben2010theory, mansour2009domain, zhao2019learning}, through a bound of the target risk:
\begin{theorem}[From \cite{ben2007analysis} and \cite{ben2010theory}] \label{ben_david} Given a hypothesis class $\mathcal H$ and a hypothesis $h\in\mathcal H$: 
\begin{equation}
    \label{BD_H}
    \varepsilon^t (h) \leq\varepsilon^s(h) + d_{\mathcal H \Delta\mathcal H} + \lambda_{\mathcal H}
\end{equation}
where $d_{\mathcal H \Delta\mathcal H} := \sup_{h,h' \in \mathcal H} | \varepsilon^s(h,h') - \varepsilon^t(h,h')|$ and $\lambda_{\mathcal H} := \inf_{h\in\mathcal H} \{ \varepsilon^t(h) + \varepsilon^s(h)\}$. In particular, provided a representation $\varphi$, and applying the inequality to $\mathcal G\circ\varphi:=\{g \circ \varphi: g \in\mathcal G\}$:
\begin{equation}
    \label{BD_G}
    \varepsilon^t(g\varphi) \leq \varepsilon^s(g\varphi) + d_{\mathcal G \Delta \mathcal G}(\varphi) + \lambda_{\mathcal G}(\varphi)
\end{equation} 
where $d_{\mathcal G \Delta \mathcal G}(\varphi) := \sup_{g,g' \in\mathcal G} |\varepsilon^s(g\varphi) - \varepsilon^t(g'\varphi)|$ and $\lambda_{\mathcal G}(\varphi) := \inf_{g \in \mathcal G} \{ \varepsilon^s(g\varphi) + \varepsilon^t(g\varphi) \}$.
\end{theorem}
On the one hand, \cref{BD_H} shows the role of the hypothesis class capacity for bounding the target risk. The lower the hypothesis class sensitivity to changes in input distribution, the lower $d_{\mathcal H \Delta \mathcal H}$. On the other hand, \cref{BD_G} puts emphasis on representations: if source and target representations are aligned \ie $p(\mathbf z^s) \approx q(\mathbf z^t)$ for $\mathbf z := \varphi(\mathbf x)$, then $d_{\mathcal G \Delta \mathcal G}(\varphi) = 0$.

\begin{figure*}
  \centering
  \subfigure[Trajectory]{
    \label{fig:trajectory}
    \includegraphics[width=.15\textwidth]{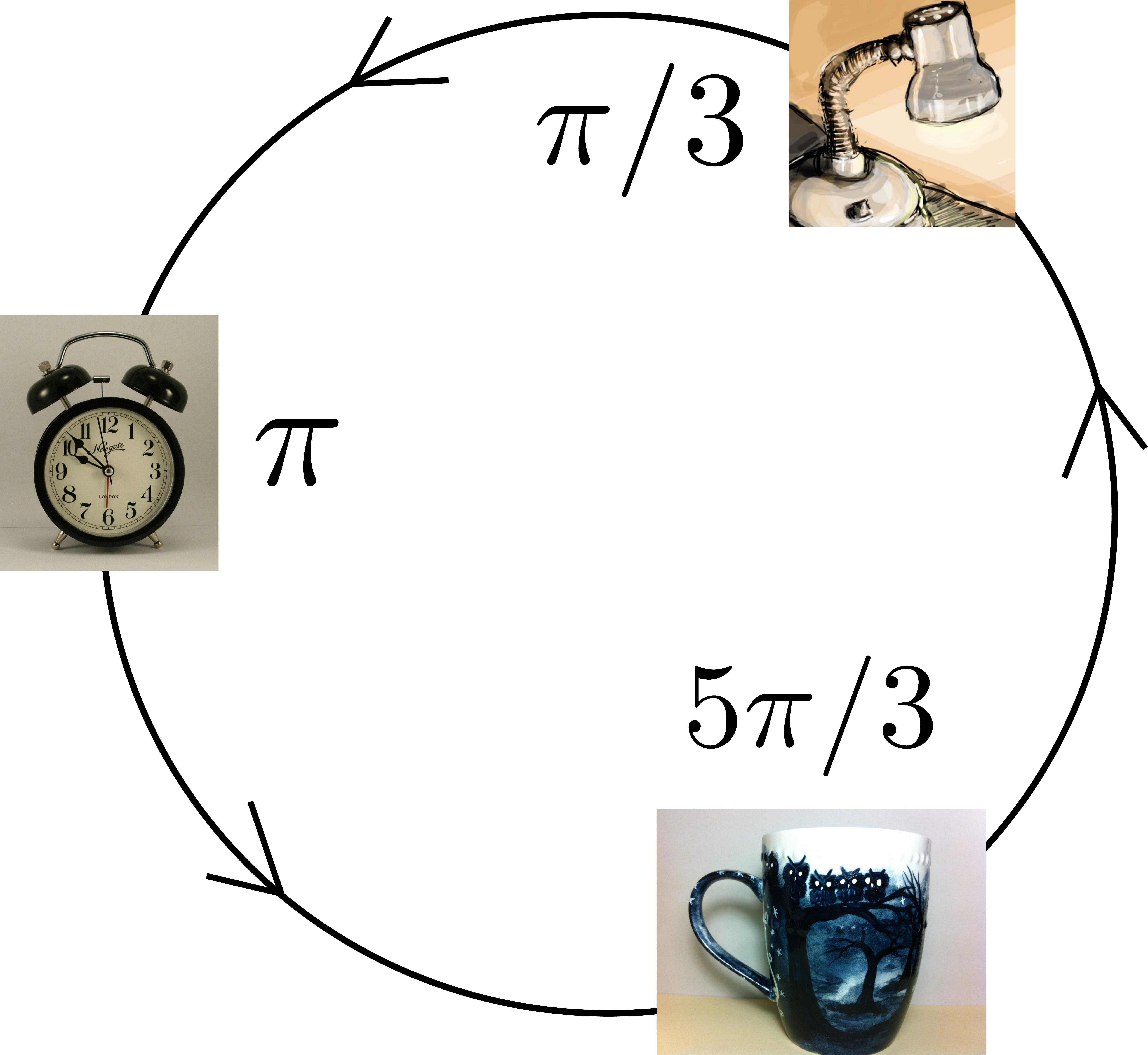}
  }
  \subfigure[Regions of Sensitivity ]{
    \label{fig:regionssensitivity}
    \includegraphics[width=.28\textwidth]{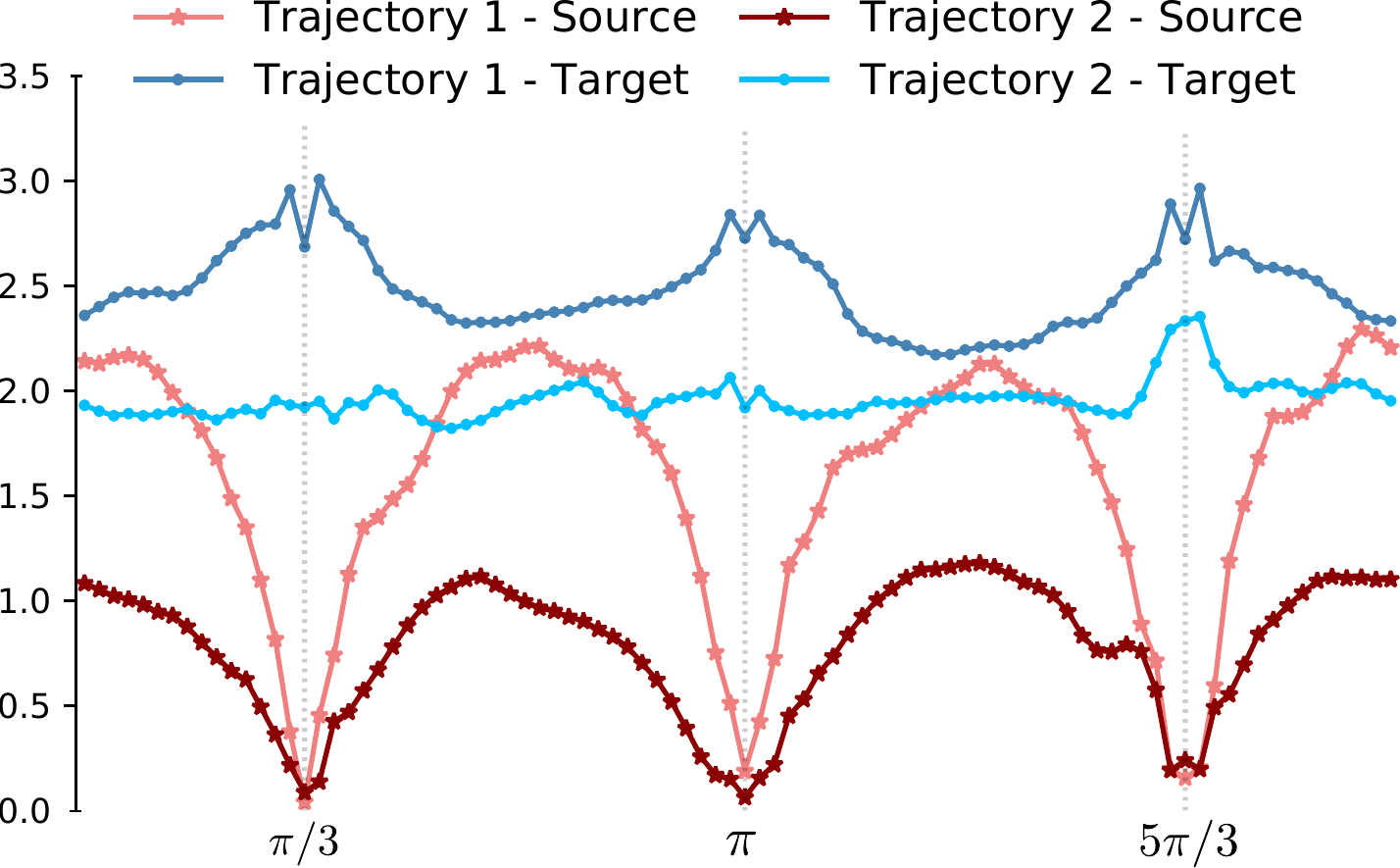}
  }
  \subfigure[Source only]{
    \label{fig:sensitivity_source}
    \includegraphics[width=.23\textwidth]{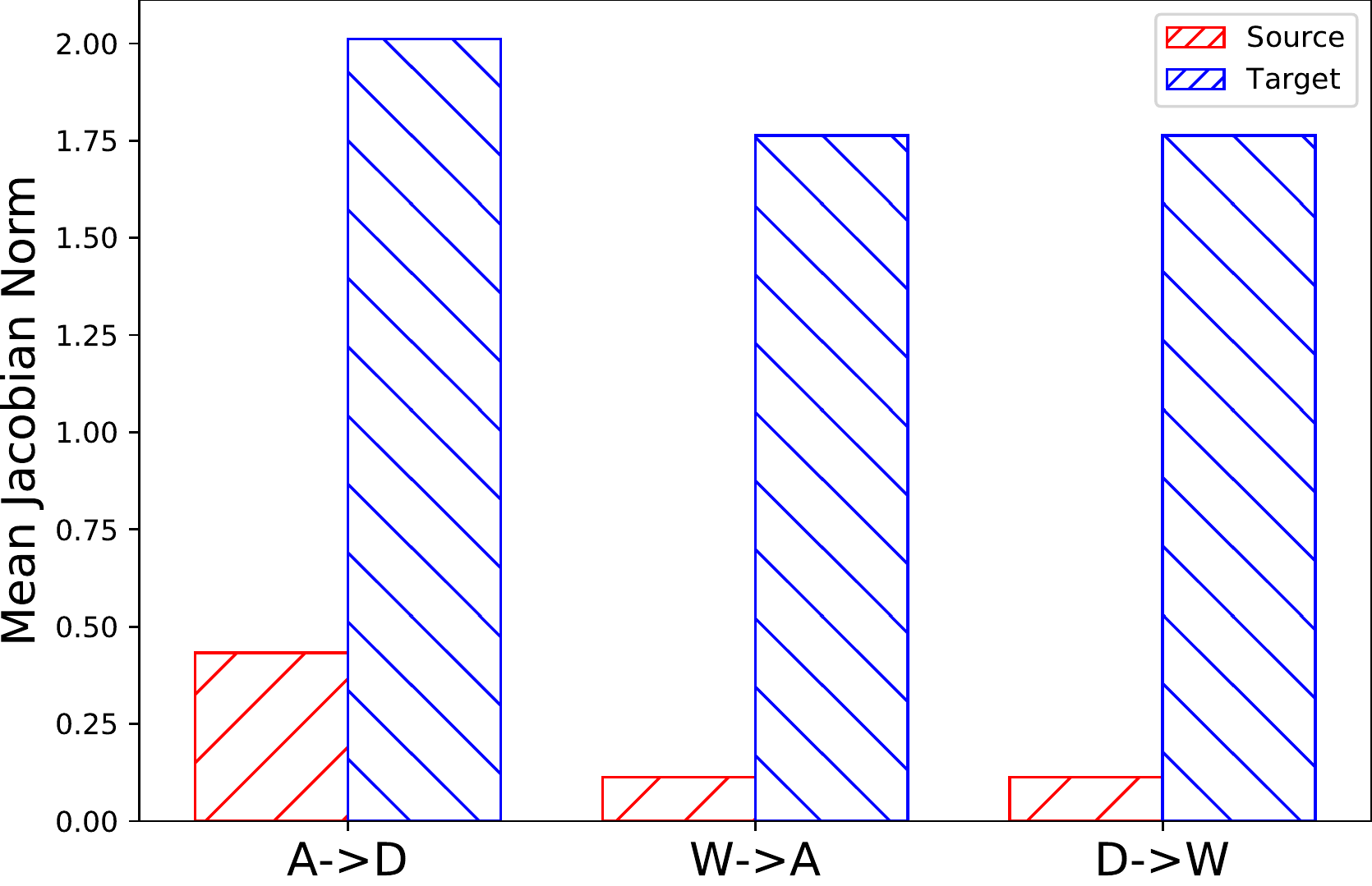}
  }
  \subfigure[DANN]{
    \label{fig:sensitivity_target}
    \includegraphics[width=.23\textwidth]{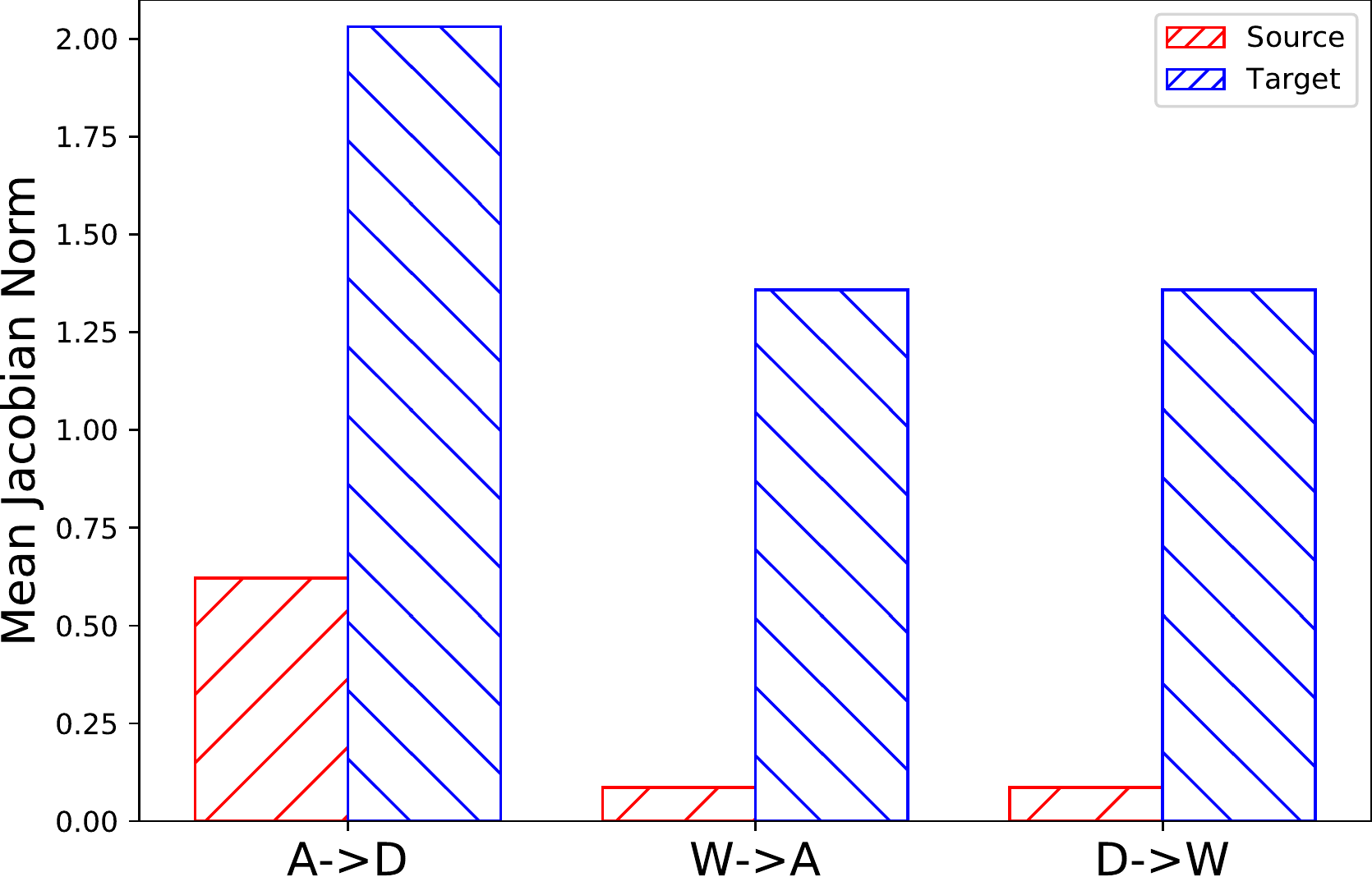}
  }
  \caption{Sensitivity Analysis. (a) An illustration of the circular trajectory passing through three images of different classes. (b) Jacobian norm of source (\textbf{D}) and target (\textbf{A}) as the input traverses two elliptical trajectories: \textit{Trajectory 1}: different classes. \textit{Trajectory 2}: same classes, for a ResNet-50 trained on source only. (c) and (d) The mean Jacobian norm on target and source domains of a ResNet-50 when trained on source only and with a DANN objective on three Office-31 tasks.}
  \vspace{-0.2in}
  \label{fig:sensitivity}
\end{figure*}

\paragraph{Sensitivity in the Target Domain.} Prior works \cite{ganin2015unsupervised, ganin2016domain, long2015learning, long2016unsupervised, long2017deep, long2018conditional} have greatly improved capacity to achieve a trade-off between source classification error and domain invariance of representations by minimizing $\varepsilon^s(g\varphi) + d_{\mathcal G \Delta \mathcal G}(\varphi)$ from \cref{BD_G}. Clearly, maintaining a low $\lambda_{\mathcal G}(\varphi)$ while learning domain invariant representations is a key to success. Some works bring theoretical evidence of its difficulty \cite{zhao2019learning, wu2019domain, johansson2019support} while pioneering works dig into that direction \cite{liu2019transferable, chen2019transferability, wang2019transferable}. This difficulty is referred as \textit{non-conservative} DA in \cite{shu2018dirt} \ie when the optimal joint classifier is significantly different from the target optimal classifier:
\begin{equation}
\inf_{h \in \mathcal{H}} \varepsilon^t(h)< \varepsilon^t(h^\lambda) \text { where } h^\lambda:=\underset{h \in \mathcal{H}}{\arg \min }\ \varepsilon^s(h)+\varepsilon^t(h)
\label{eq:optimality_gap}
\end{equation}
Similarly, when provided with a representation $\varphi$, the optimal joint classifier differs from the target optimal classifier: $\inf_{g\in\mathcal G} \varepsilon^t(g\varphi) < \varepsilon^t(g^\lambda\varphi)$ where $g^\lambda := \arg \min_{g \in \mathcal G}\{ \varepsilon^s(g\varphi) + \varepsilon^t(g \varphi) \}$ (see Appendix for more details). %\footnote{\cite{shu2018dirt} describes non-conservative DA 
%from the hypothesis class point of view, \ie \cref{BD_H} from \cref{ben_david}, then allowing change in representations to detect it, \ie $\inf$ computed on $\mathcal H$ in \cref{eq:optimality_gap}. The literature of domain adversarial learning puts emphasis on representations, \ie \cref{BD_G} from theorem \cref{ben_david}, and the current definition only allows to modify the classifier, $\inf$ computed on $\mathcal G$, for detecting non-conservative DA. We extend the denomination of non conservative DA to the case where $\varepsilon^t(\varphi):=\inf_{g\in\mathcal G} \varepsilon^t(g\varphi)$ is not optimal in $\varphi$.}. 
Importantly, mitigating at train time the risk of non-conservative DA is a difficult problem since target labels are involved in \cref{eq:optimality_gap}. Therefore, other tools need to be leveraged to detect non-conservative adaptation without the ground truth in the target domain. Following the insight from \cite{shu2018dirt}, we hypothesize that violation of the cluster assumption in the target domain is a strong indicator of a case of non-conservative DA. In such a case, a classifier with different source and target errors should exhibit a substantial sensitivity in the target domain to small input perturbations. Indeed, a violation of the cluster assumption is characterized by a decision boundary localized in high density regions of the target input space. 

%If this hypothesis is verified, a classifier with different source and target errors should exhibit a substantial sensitivity in the target domain since the violation of the cluster assumption is characterized by a decision boundary localized in a high density region of the target input space. 

Therefore, we analyze the robustness of a model trained to minimize the source risk, through its sensitivity to small perturbations in the input space. %distinguish different inputs in the neighborhood of target data points. A model which verifies the cluster assumption in the target domain is expected to have consistent outputs over small input perturbations. Then, we study the sensitivity of the model to small perturbation of the inputs. 
%The sensitivity of the model in the target domain can be estimated by answering \textit{how does the output of the model change as the input target example is perturbed?} 
We follow \cite{NovakBAPS18} and compute the mean Jacobian norm as a proxy of the generalization at the level of individual target sample, and as a measure of the local sensitivity of the model on target examples: $\mathbb{E}_{\mathbf x^t \sim q}\left [\left\| \mathsf J \left(\mathbf  x^t\right )\right \|_{F}\right]$ where  $\mathsf J_{i j}(\mathbf{x})=\partial \hat{y}_{i} / \partial x_{j}$ is the Jacobian matrix,
%\begin{equation}
%\mathbb{E}_{\mathbf x^t \sim q}\left [\left\| \mathsf J \left(\mathbf  x^t\right )\right \|_{F}\right] ~\text{ where } ~ \mathsf{J}(\mathbf{x})= \left (\frac{\partial \hat{\mathbf y}}{\partial \mathbf{x}}\right)^\top
%\label{eq:jacobianmean}
%\end{equation}
%where 
 $\| \mathsf J \|_{F}$ is the Frobenius norm, and $\hat{y}_{i}$ is the output class probability for class $i$. For comparison, the sensitivity in the source domain can be computed in a similar manner over source instances. By language abuse, we will refer to sensitivity in source and target domains as source and target sensitivity, respectively. The results obtained on 3 transfer tasks from Office-31 (\textbf{A} $\rightarrow$ \textbf{D}, \textbf{W} $\rightarrow$ \textbf{A}, \textbf{D} $\rightarrow$ \textbf{W}) are shown in \cref{fig:sensitivity_source} and \cref{fig:sensitivity_target}. As suspected, the target sensitivity is significantly higher compared to the source sensitivity. Importantly, when enforcing invariance of representations with Domain Adversarial Neural Networks (DANN \cite{ganin2015unsupervised}), sensitivity in the target domain decreases (for tasks  \textbf{W} $\rightarrow$ \textbf{A} and \textbf{D} $\rightarrow$ \textbf{W}) while remaining significantly higher than the source sensitivity. This validates our concern on non-conservative domain adaptation: even after features alignment, the resulting classifier still violates the cluster assumption in the target domain. To further investigate the regions of sensitivity, we examine the behavior of the function on and off the data manifold as it approaches and moves away from three anchor points. To this end, following \cite{NovakBAPS18}, we analyze the behavior of the model near and away from target and source data along two types of trajectories: 1) an ellipse passing through three data points of different classes as illustrated in \cref{fig:trajectory}, and 2) an ellipse passing through three data points of the same class.
Since linear combinations of images of the same class are likely to look like a realistic image, the second trajectory is expected to traverse overall closer to the data manifold.
\cref{fig:regionssensitivity} shows the obtained results. We observe that, according to the Jacobian norm, the model's sensitivity in the vicinity of target data is comparable to its sensitivity off the data manifold. Inversely, the model remains relatively stable in the neighborhood of source data and becomes unstable only away from them, further confirming our hypothesis.

\section{Target Consistency}

\subsection{Consistency Regularization}
In order to promote a more robust model and mitigate target sensitivity, we regularize the model predictions to be invariant to a set of perturbations applied to the target inputs. %, since a model robust
%to any small changes in the input space naturally complies with such an assumption. 
Concretely, we add to the objective function an additional Target Consistency term:
\begin{equation}
	\begin{split}
	\label{eq:target_consistency}
	\mathcal L_{\mathrm{TC}}(\varphi, g) &= \mathcal L_{\mathrm{VAT}}(\varphi, g) + \mathcal L_{\mathrm{AUG}}(\varphi, g)\\
	&= \mathbb E_{\mathbf x^t \sim p} \left[\max _{\|r\| \leq \epsilon} ||(h(\mathbf x^t) - h(\mathbf x^t + r)||^2\right]
	+ \mathbb E_{\mathbf x^t \sim p} \left[||(h(\mathbf x^t) - h(\mathbf{\tilde{x}}^t)||^2\right]
	\end{split}
\end{equation}
Similar to \cite{shu2018dirt}, the first term incorporates the locally-Lipschitz constraint by applying Virtual Adversarial Training (VAT) \cite{miyato2018virtual} which forces the model to be consistent within the norm-ball neighborhood of each target sample $\mathbf x^t$. Additionally, the second term forces the model to embed a target instance $\mathbf x^t$ and its augmented version $\mathbf{\tilde{x}}^t$ similarly to push for smooth neural network responses in the vicinity of each target data. With a carefully chosen set of augmentations, such a constraint makes sense since the semantic content of a transformed image is approximately preserved. Note that for more stable training, we follow Mean Teachers (MT) \cite{tarvainen2017mean} and an use of an exponential moving average of the model to compute the target pseudo-labels (\ie $h(\mathbf x^t)$). Overall, $\mathcal L_{\mathrm{TC}}$ is in-line with the cluster assumption by promoting consistency to various set of input perturbations, thus, forcing the decision boundary to not cross high-density regions.

\begin{figure}[tbp]
  \centering
  \includegraphics[width=0.9\textwidth]{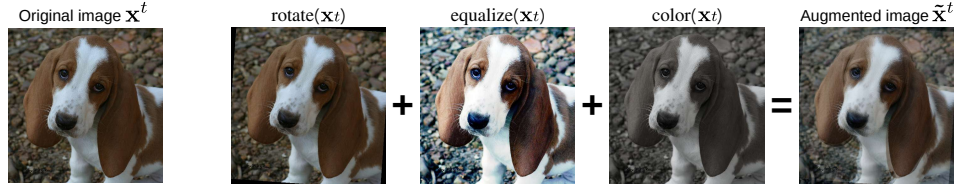}
  \caption{Mixing Augmentations. An example of an augmented image with $K = 3$. We start by sampling 3 operations, rotate, equalize and color, which are then applied to the original image. The augmented image can then be obtained using an element-wise convex combination of the augmented images, resulting in a semantically similar image, while still injecting a higher degree of noise.}
  \vspace{-0.15in}
  \label{fig:augs}
\end{figure}

\textbf{Augmentations.} For visual domain adaptation, and based on the recent success of supervised image augmentations
\cite{cubuk2019autoaugment,lim2019fast,cubuk2019randaugment} in semi-supervised learning \cite{xie2019unsupervised,berthelot2019remixmatch,sohn2020fixmatch}
and robust deep learning \cite{yin2019fourier,hendrycks2019augmix}, we propose to use a rich set of state-of-the-art data augmentations to inject noise and enforce a consistency of predictions on target domain. Specifically, we use augmentations from AutoAugment \cite{cubuk2019autoaugment}. Upon each application, we sample a given operation $o$ from all possible augmentations $\mathcal O = \{\text{equalize}, \ldots, \text{brightness}\}$. If the operation $o$ is applicable with varying severities, we also uniformly sample the severity, and apply $o$ to obtain the augmented target image $\mathbf{\tilde{x}}^t = o(\mathbf x^t)$. However, applying a single operation might be solved easily by a high capacity model by memorizing the specific perturbations. To overcome this, we generate more diverse augmentations by mixing multiple augmented images (see \cref{fig:augs}). We start
by randomly sampling $K$ operations from $\mathcal O$ and $K$ convex coefficients $\alpha_i$ sampled from a Dirichlet distribution: $(\alpha_1, \ldots, \alpha_K) \sim \text{Dir} (1, \ldots, 1)$. The augmented image $\mathbf{\tilde{x}}^t$ can then be obtained with an element-wise convex combination of the $K$ augmented instances of $\mathbf x^t$: $\mathbf{\tilde{x}}^t = \sum_{i=1}^K \alpha_i o_i(\mathbf x^t)$, impelling the model to be stable, consistent, and insensitive across a more diverse range of inputs \cite{zheng2016improving,kannan2018adversarial,hendrycks2019augmix}.

\subsection{Coupling Target Consistency with Invariant Representations}
\begin{figure}[]
    \centering
    \includegraphics[width=1.\textwidth]{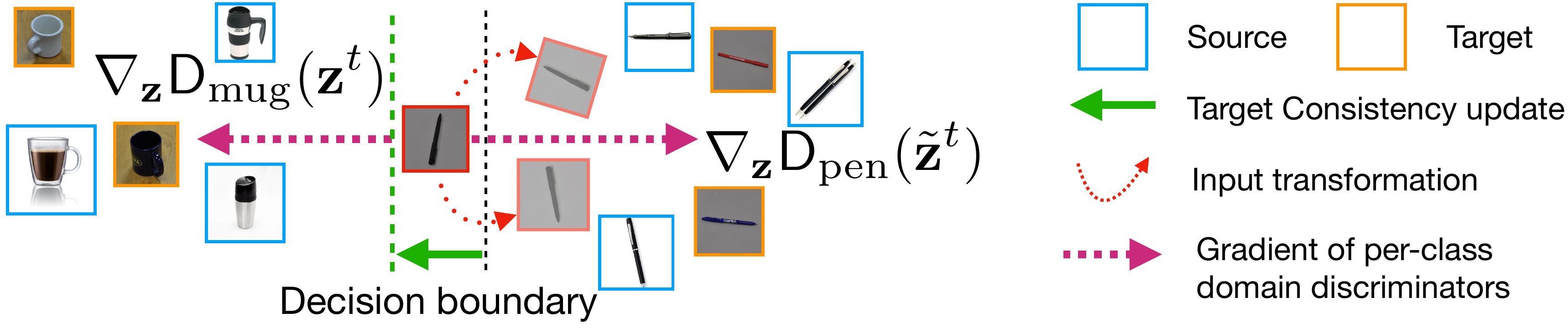}
    \caption{Effect of TC on the learned representations. Mugs and pens from the source (\textbf{A}) and target (\textbf{D}) domains of Office31 are pictured. The red squared pen, a target sample, is confounded with a mug due to spurious correlations \ie upward orientation and black color. Input augmentations wipe out spurious correlations induced by the orientation, and the TC pushes the decision boundary to low density regions, correcting the predicted class. Before the TC update, the class-level discriminator encourages the pen to reach the high-density region of the mug class. At this time, the class-level discriminator and TC gradients have opposite directions, indicating a negative interaction. The TC update allows the sample to cross the decision boundary. It ultimately changes the class-level discriminator which now encourages to reach high-density region of the pen-class. At this time, the domain adversarial and TC gradients have similar directions, indicating a positive interaction. Crucially, the gradient of a vanilla domain discriminator (DANN \cite{ganin2015unsupervised}) interacts poorly with the TC update since it does not modify the target representations distribution. \textit{Best viewed in color.}}
    \vspace{-0.15in}
    \label{fig:TC_IR}
\end{figure}

\textbf{Target Consistency and $d_{\mathcal H \Delta \mathcal H}$.} Enforcing the target consistency gives us the ability to control the trade-off between a low target sensitivity, \ie a low violation of the cluster assumption, and a low source risk. As described in \cite{shu2018dirt}, adding $\mathcal L_{\mathrm{TC}}$ to the objective function reduces the hypothesis class $\mathcal H$ to only include classifiers that are robust on both target and source domains, noted $\mathcal H_{\mathrm{TC}}$. Through the lenses of domain adaptation theory (\ie \cref{BD_H} from \cref{ben_david}), by constraining the hypothesis space $\mathcal H$ to contain stable classifiers across domains, small changes to the hypothesis in the source domain will not induce large changes in the target domain \cite{shu2018dirt}. Formally, this reduces the domain discrepancy $ d_{\mathcal{H}_{\mathrm{TC}} \Delta \mathcal{H}_{\mathrm{TC}}}  \leq d_{\mathcal{H} \Delta \mathcal{H}}$ based on the following inclusion $\mathcal H_{\mathrm{TC}} \subset \mathcal H$. 

%First, we consider target samples near the decision boundary since such instances are hard to adapt and are responsible for reducing the classifier generalization on target. Injecting a small perturbation to such low-margin target samples $x^t$ with likely push their perturbed versions, $\tilde{x}$ to the other side of the decision boundary, violating the cluster assumption. By enforcing TC, we incrementally and locally push the decision boundary as far as possible from the class boundaries, obtaining consistent, and hopefully correct predictions, on low-margin target samples. However, even if such incremental updates on low-margin target samples might result in correcting the predicted class label, the underlining representations remain approximately the same, and the discriminator feedback does not reflect this predicted labels change. 

However, viewing the effect of TC as a constraint on the hypothesis space (\cref{BD_H} from \cref{ben_david}) does not explain the hidden interactions between TC and invariant representations  (\cref{BD_G} from \cref{ben_david}).
To this purpose, we consider a target sample $\mathbf x^t$ near the decision boundary (red squared pen in \cref{fig:TC_IR}) which is hard to adapt (the pen is confounded with a mug). Thus, its augmented version, $\tilde{\mathbf x}^t$, is likely to have a different predicted class (red squared pens with a low opacity in \cref{fig:TC_IR}). By enforcing TC, the model embeds $\mathbf x^t$ and $\tilde{\mathbf x}^t$ similarly to incrementally push the decision boundary far from class boundaries. Such incremental change might result in correcting the predicted class label (green decision boundary in \cref{fig:TC_IR}). However, the underlining representations remain approximately the same, and the discriminator feedback does not reflect this predicted labels change. Now, consider that domain invariance is achieved by leveraging one discriminator per predicted class \ie class-level invariance. The change of label due to the TC update will result in a switch of the discriminator used, subsequently reflecting the label change in the domain adversarial loss. This interaction between class-level invariance and decision boundary update is the key to the success of TC.

\textbf{Class-level domain discriminator.} We introduce CLIV, a well-suited Class-Level InVariance adversarial loss, which leverages one discriminator per predicted class. Let  $\mathsf D := (D_c)_{1 \leq c \leq C}$, a set of $C$ discriminators \ie for $\mathbf z \in \mathcal Z,\mathsf D(\mathbf z) \in [0,1]^C$. Given a sample $\mathbf x$ with representation $\mathbf z$ and output $\hat{\mathbf y} := g(\mathbf z)$,  we weight the importance of discriminator $D_c$ in the adversarial loss using the ouptut $\hat{\mathbf y}$. This results into a class conditioning of the domain adversarial loss, where the ground-truth is used in the source domain and noting $\cdot$ the scalar product in $\mathbb R^C$:
\begin{equation}
    \mathcal L_{\mathrm{CLIV}}(\varphi) := \inf_{\mathsf D} \left \{ \mathbb E_{(\mathbf z^s, \mathbf y^s) \sim p} \left [\mathbf y^s \cdot \log (\mathsf D \left (\mathbf z^s) \right ) \right ] + \mathbb E_{(\mathbf z^t, \hat{\mathbf y}^t) \sim q}\left [\hat{\mathbf y}^t \cdot \log \left (1 - \mathsf D(\mathbf z^t) \right) \right ] \right \}
\end{equation}
Our model is trained by minimizing a trade-off between source Cross-Entropy (CE),
Class-Level InVariance (CLIV) and Target Consistency (TC); given $\lambda_{\mathrm{CLIV}}$ and $\lambda_{\mathrm{TC}}$ tunable hyper-parameters,
\begin{equation}
    \mathcal L(g,\varphi) := \mathcal L_{\mathrm{CE}}(g,\varphi) + \lambda_{\mathrm{CLIV}} \mathcal L_{\mathrm{CLIV}}(\varphi) + \lambda_{\mathrm{TC}}  \mathcal L_{\mathrm{TC}}(g,\varphi) 
\end{equation}

\textbf{Theoretical analysis.} We provide theoretical insights into the interaction between TC and class-level invariance. We consider $\varphi \in \Phi$ and $g \in \mathcal G$, which are modified to obtain $\tilde \varphi$ and $\tilde g$ defined as the closest instances such that $\tilde g\tilde \varphi$ verifies TC. For instance, they can be obtained by minimizing  $ \ell_2(\varphi, \tilde \varphi)  +  \ell_2( g , \tilde g)  +  \lambda \cdot \mathcal L_{\mathrm{TC}}(\tilde g, \tilde \varphi)$ where $\ell_2$ is an $L^2$ error. When enforcing TC, we expect to decrease the target error \ie $
    \varepsilon^t (\tilde g \tilde \varphi) < \varepsilon^t(g\varphi)  $. Noting $\rho := (1- \varepsilon^t (\tilde g \tilde \varphi) / \varepsilon^t(g\varphi))^{-1}$ and $\tilde{\mathbf y} := \tilde g \tilde\varphi(\mathbf x)$, $\mathsf F$ a large enough critic function space (See Appendix), we adapt the theoretical analysis from \cite{bouvier2020robust}: 
\begin{equation}
    \label{bound_final}
    \varepsilon^t( g \varphi) \leq \rho \left ( \varepsilon^s(g\varphi) + 8\sup_{\mathsf f \in \mathcal F} \left \{ \mathbb E_{(\mathbf z^s, \mathbf y^s) \sim p}[ \mathbf y^s \cdot \mathsf f(\mathbf z^s)] - \mathbb E_{(\mathbf z, \tilde{\mathbf y}) \sim q}[ \tilde{ \mathbf y}^t \cdot \mathsf f(\mathbf z^t)] \right\}  + \inf_{\mathsf f \in \mathcal F} \varepsilon^t(\mathsf f \varphi) \right )
\end{equation}
Crucially, by observing that $\sup_{\mathsf f \in \mathcal F} \left \{ \mathbb E_{(\mathbf z^s, \mathbf y^s) \sim p}[ \mathbf y^s \cdot \mathsf f(\mathbf z^s)] - \mathbb E_{(\mathbf z, \tilde{\mathbf y}) \sim q}[ \tilde{ \mathbf y}^t \cdot \mathsf f(\mathbf z^t)] \right\}$ is an \textit{Integral Probability Measure} proxy of $\mathcal L_{\mathrm{CLIV}}$, \cref{bound_final} reveals that class-level domain invariant representations can leverage feedback from an additional regularization, here the Target Consistency, to learn more transferable invariant representations.

\section{Experiments}
%In this section, we evaluate the proposed method on two major visual domain adaption recognition tasks: classification and segmentation.

\subsection{Setup}
\textbf{Datasets.} \textbf{Office-31} \cite{saenko2010adapting} is the standard dataset for visual domain adaptation, containing 4,652 images in 31 categories divided across three domains:
Amazon (\textbf{A}), Webcam (\textbf{W}), and DSLR (\textbf{D}).
We use all six possible transfer tasks to evaluate our model. \textbf{ImageCLEF-DA} \href{https://www.imageclef.org/2014/adaptation}{$^*$} %\footnote{\url{https://www.imageclef.org/2014/adaptation}}  
 is a dataset with 12 classes and 2,400 images assembled from three public datasets: Caltech-256 (\textbf{C}), ImageNet (\textbf{I}) and Pascal VOC 2012 (\textbf{P}), where each one is considered as separate domain. We evaluate on all possible pairs of the three domains. \textbf{Office-Home} \cite{venkateswara2017deep} is a more difficult dataset compared to Office-31, consisting of 15,00 images across 65 classes in office and home settings. The dataset consists of four widely different domains: Artistic images (\textbf{Ar}), Clip Art (\textbf{Ca}), Product images (\textbf{Pr}), and Real-World images (\textbf{Rw}). We conduct experiments on all twelve transfer tasks. \textbf{VisDA-2017} \cite{peng2017visda} presents a challenging simulation-to-real dataset, with two very distinct domains: \textbf{Synthetic}, with renderings of 3D models with different lightning conditions and from many angles; \textbf{Real} containing real-world images. We conduct evaluations on the \textbf{Synthetic} $\rightarrow$ \textbf{Real} task. For semantic segmentation experiments, we evaluate our method on the challenging \textbf{GTA5} $\rightarrow$ \textbf{Cityscapes} VisDA-2017 semantic segmentation task. The synthetic source domain is \textbf{GTA5} \cite{richter2016playing} dataset with 24,966 labeled images, while the real target domain is \textbf{Cityscapes} \cite{cordts2016cityscapes} dataset consisting of 5,000 images. Both datasets are evaluated on the same set of 19 classes, with the mean Intersection-over-Union (mIoU) metric.

\textbf{Protocol.} We follow the standard protocols for UDA \cite{long2017deep,long2018conditional,chen2017deeplab}. We train on all labeled source samples and all unlabeled target samples and compare the %average 
classification accuracy based on three random experiments for classification and the mIoU based on a single run for segmentation. 
For classification, we use the same hyperparameters as CDAN \cite{long2018conditional} and adopt ResNet-50 \cite{he2016deep} as a base network pre-trained on ImageNet dataset \cite{deng2009imagenet}.
As for CLIV and TC hyperparameters, we use $K = 4$, $\lambda_{\mathrm{CLIV}} = 1$ and $\lambda_{\mathrm{TC}} = 10$. We note that the method performs comparatively on a wide range of hyperparameter values making it robust for practical applications.
For segmentation, we follow ADVENT \cite{vu2019advent} and use the same experimental setup with Deeplab-V2 \cite{chen2017deeplab} as the base semantic segmentation architecture with a ResNet-101 backbone and a DCGAN discriminator \cite{radford2015unsupervised}. We employ \texttt{PyTorch} \cite{paszke2019pytorch} and base our code on official implementations of CDAN \cite{long2018conditional} and ADVEN \cite{vu2019advent}. 

\subsection{Results}

\begin{table*}[htbp]
  \vspace{-12pt}
  \addtolength{\tabcolsep}{2pt}
  \centering
  \caption{Average accuracy (\%) of all tasks on image classification benchmarks for UDA. We compare our approach with similar methods based on invariant representations, evaluated using the same protocol. Results are obtained with a ResNet-50 unless specified otherwise. For detailed per task results see the Appendix.}
  \label{table:classification}
  \resizebox{\textwidth}{!}{%
  \begin{tabular}{lccccccc}
    \toprule
    Method & Office-31 & ImageCLEF-DA & Office-Home & VisDA & VisDA (ResNet-101)\\
    \midrule
    ResNet & 76.1 & 80.7 & 46.1 & 45.6 & 52.4 \\
    %DAN \cite{long2015learning} & 80.4 & 82.5 & 56.3 \\
    %RTN \cite{long2016unsupervised} &  81.6 & - & - \\
    DANN \cite{ganin2016domain} &  82.2 & 85.0 & 57.6 & 55.0 & 57.4  \\
    %ADDA \cite{tzeng2017adversarial}& 82.9 & - &  56.3 \\
    %JAN \cite{long2017deep} & 84.3 & 85.8 & 58.3 \\
    %GTA \cite{sankaranarayanan2018generate} & 86.5 & - & - \\
    CDAN \cite{long2018conditional} & 87.7 & 87.7 & 65.8 & 70.0 & 73.7\\
    TAT \cite{liu2019transferable} & 88.4 & 88.9 & 65.8 & 71.9 & - \\
    BSP \cite{chen2019transferability} & 88.5 & - & 66.3 & - & 75.9 \\
    TransNorm \cite{wang2019transferable} & 89.3 & 88.5 & 67.6 & 71.4 & - \\
    \textbf{Ours} & \textbf{89.6} & \textbf{89.5} & \textbf{69.0} & \textbf{77.5} & \textbf{79.0}\\    
    \bottomrule
  \end{tabular}
  }
  \vspace{-5pt}
\end{table*}

%For clarity and compactness, the results of the average accuracy of all the tasks on all of the standard classification benchmarks for UDA are reported in \cref{table:classification}. The detailed per task results can be found in the Appendix. The proposed method outperforms previous adversarial methods on all datasets. The gains are substantial when the source and target domain are more dissimilar as in VisDA dataset, we conjuncture that this is a result of the large number of target instances available, enabling us to extract a significant amount of training signal with TC objective term to enforce the cluster assumption. Additionally, the method performs well with a large number of categories, as it is the case for Office-Home dataset, where the class level discrimination helps alleviate the difficulty of the adaption. We observe overall smaller improvements on Office-31 due to its limited size, and ImageCLEF-DA which is expected since the three domains are visually more similar, making the task of domain adaptation easier. Our method also performs well compared to non-adversarial methods, such as DTA \cite{lee2019drop}, as shown in \cref{table:visda}.

For clarity and compactness, the results of the average accuracy of all the tasks on all of the standard classification benchmarks for UDA are reported in \cref{table:classification}. The proposed method outperforms previous adversarial methods on all datasets. The gains are substantial when the source and target domain are more dissimilar as in VisDA dataset. We conjuncture that this is a result of a large number of target instances available, enabling us to extract a significant amount of training signal with TC objective term to enforce the cluster assumption. Additionally, the method performs well with a large number of categories, as it is the case for Office-Home dataset. Such gain is a result of the class-level invariance which is empowered as the number of classes grows. We observe overall smaller improvements on Office-31 due to its limited size, and ImageCLEF-DA since the three domains are visually more similar. %Our method also performs well compared to non-adversarial methods, such as DTA \cite{lee2019drop}, as shown in \cref{table:visda}. 
We further demonstrate the generality of the proposed method by conducting additional experiments on GTA5 $\shortrightarrow$ Cityscapes task for semantic segmentation (\cref{table:segmentation}), and observe a gain of 2.5 points over the baseline Adapt-SegMap \cite{tsai2018learning}, confirming the flexibility of TC and its applicability across DA tasks.

\begin{table}[!htbp]
\vspace{-10pt}
\addtolength{\tabcolsep}{6pt}
\centering 
\begin{minipage}{0.4\textwidth}
  \centering
  \caption{mIoU on \textbf{GTA5 $\shortrightarrow$ Cityscapes}. 
  {\scriptsize{\textit{AdvEnt+MinEnt}* is an ensemble of two models.}}}
	\label{table:segmentation}
  	\resizebox{0.8\textwidth}{!}{%
  \begin{tabular}{lc}
  	\toprule
  	\multicolumn{2}{c}{DeepLab v2}\\
    Method & mIoU \\
    \midrule
    Adapt-SegMap \cite{tsai2018learning} & 42.4 \\
    AdvEnt \cite{vu2019advent} & 43.8 \\
    \textbf{Ours} & \textbf{44.9} \\
    \midrule
    AdvEnt+MinEnt* \cite{vu2019advent} & \underline{\textbf{45.5}} \\
    \bottomrule
  \end{tabular}
 }
  \vspace{-0.2in}
\end{minipage}%
\hfill
\begin{minipage}{0.57\textwidth}
  \centering
  \vspace{0.25in}
   \caption{Avg Acc (\%) of the 5 hardest Office-Home tasks for TC coupled with different adversarial losses.}
   	\label{table:ablation_Ladv}
   	\resizebox{1\textwidth}{!}{%
      \begin{tabular}{lccc}
        \toprule
        $\mathcal L_{\mathrm{adv}} =$ & $\mathcal L_{\mathrm{DANN}}$ \cite{ganin2015unsupervised} & $\mathcal \mathcal L_{\mathrm{CDAN}}$ \cite{long2018conditional} & $ \mathcal L_{\mathrm{CLIV}}$ \\
        \midrule
        $\mathcal L_{\mathrm{adv}}$  & 47.6 & 53.4 & 56.7 \\
        $~ + \mathcal L_{\mathrm{VAT}}$  & 48.0 & 55.1 & 57.1  \\
        $~  + \mathcal L_{\mathrm{AUG}}$  &  51.3 & 55.7 & 58.1 \\
        $~ + \mathcal L_{\mathrm{VAT}}+ \mathcal L_{\mathrm{AUG}}$  & 51.4 & 56.9 & 58.6  \\
        $~ + \mathcal L_{\mathrm{VAT}}+ \mathcal L_{\mathrm{AUG}}$ w/ $\mathrm{MT}$  &  51.0 & 56.0 & \textbf{58.9} \\
        \bottomrule
      \end{tabular}
  }
  \vspace{-5pt}
\end{minipage}
\end{table}

\subsection{Ablations}

To examine the effect of each component of our proposed method, we conduct several ablations on the 5 hardest tasks on Office-Home, with and without the TC term, and with different variations of the TC loss. The results are reported in \cref{table:ablations_home}. We observe that adding a consistency term, either VAT or AUG, results in a higher accuracy across tasks, with better results when smoothing in the vicinity of each target data point within the data manifold with AUG, instead of the adversarial direction using VAT. Their combination, with Mean Teacher (MT), results in an overall more performing model.
%, does increase the results for an overall more stable model, with a modest improvement when using a Mean Teacher (MT) to produce the target pseudo-labels.

\begin{table}[!htbp]
\vspace{-5pt}
\addtolength{\tabcolsep}{6pt}
\centering 
\begin{minipage}{0.70\textwidth}
  \centering
  \caption{Acc (\%) on the 5 hardest {Office-Home} tasks for TC ablation.}
  \label{table:ablations_home}
  \resizebox{1\textwidth}{!}{%
  \begin{tabular}{lcccccc}
    \toprule
    & Ar$\shortrightarrow$Cl & Cl$\shortrightarrow$Ar & Pr$\shortrightarrow$Ar & Pr$\shortrightarrow$Cl & Rw$\shortrightarrow$Cl & Avg \\
    \midrule
    $\mathcal L_{\mathrm{CLIV}}$ & 52.6 & 60.1 & 60.6 & 52.1 & 58.3 &  56.7 \\ 
    $~ + \mathcal L_{\mathrm{VAT}}$ & 52.4 & 60.1 & 61.2 & 53.1 & 58.9 &  57.1 \\ 
    $~  + \mathcal L_{\mathrm{AUG}}$ & 53.1 & 62.3 & 62.6 & 53.1 & 59.5 & 58.1 \\ 
    $~ + \mathcal L_{\mathrm{VAT}}+ \mathcal L_{\mathrm{AUG}}$ & 53.0 & \textbf{62.8} & 62.8 & 53.8 & \textbf{60.8} &  58.6 \\ 
    $~ + \mathcal L_{\mathrm{VAT}}+ \mathcal L_{\mathrm{AUG}}$ w/ $\mathrm{MT}$ & \textbf{53.1} & 62.6 & \textbf{63.8} & \textbf{54.4} & 60.4 &  \textbf{58.9} \\ 
    \bottomrule
  \end{tabular}
 }
  %\vspace{-0.2in}
\end{minipage}%
\hfill
\begin{minipage}{0.27\textwidth}
      \includegraphics[width=0.85\textwidth]{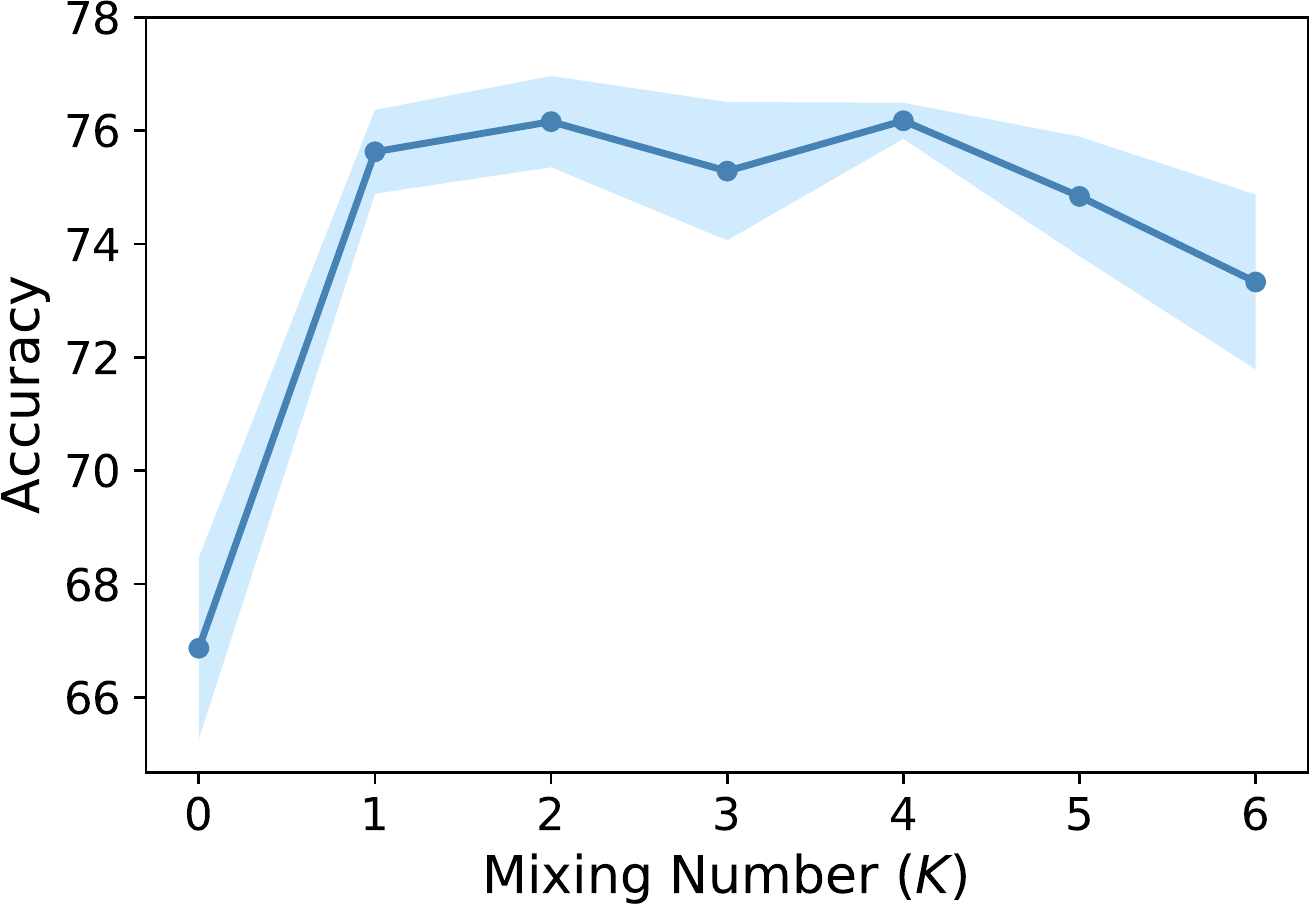}
      \captionof{figure}{Effect of mixed augmentations $K$ (VisDA).}
      %\caption{Accuracy on VisDA-2017 with different number of mixed augmentations $K$.}
      \vspace{-0.15in}
      \label{fig:ablation_augs}
  %\vspace{-5pt}
\end{minipage}
\end{table}

We also conduct an ablation study on the effect of varying the mixing number $K$ to produce more diverse target images. \cref{fig:ablation_augs} shows the results. Overall, we observe a slight improvement and more stable results when $K$ is increased, but over a certain threshold, the degree of noise becomes significant, heavily modifying the semantic content of the inputs, and hurting the model's performance.

Most importantly, to show the importance of coupling TC with CLIV, we pair TC with DANN and CDAN losses, and the obtained results in \cref{table:ablation_Ladv} show lower average accuracy and minimal gains when enforcing the cluster assumption in conjunction with such adversarial losses, confirming the importance of imposing class-level invariance when applying TC.

\subsection{Analyses}

\begin{figure}[!htbp]
\vspace{-10pt}
\centering
\begin{minipage}{0.33\textwidth}
      \includegraphics[width=0.9\textwidth]{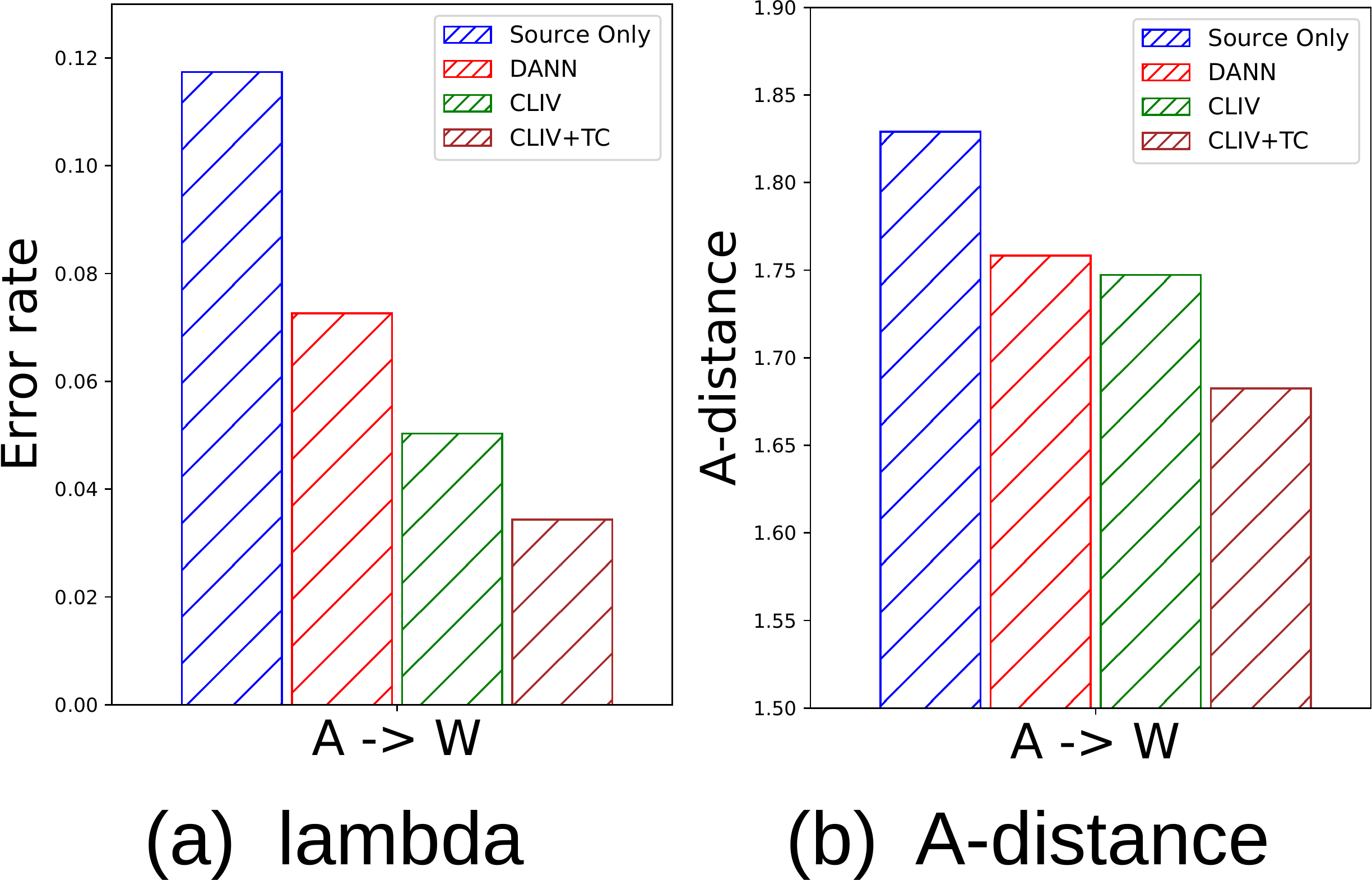}
      \caption{A-distance and $\lambda$.}
      \vspace{-0.15in}
      \label{fig:lambda_and_discrepancy}
\end{minipage}%
\hfill
\begin{minipage}{0.63\textwidth}
  \centering
  \subfigure[\textbf{A $\shortrightarrow$ W}]{
    \label{fig:sensitivity_TC_AW}
    \includegraphics[width=0.46\textwidth]{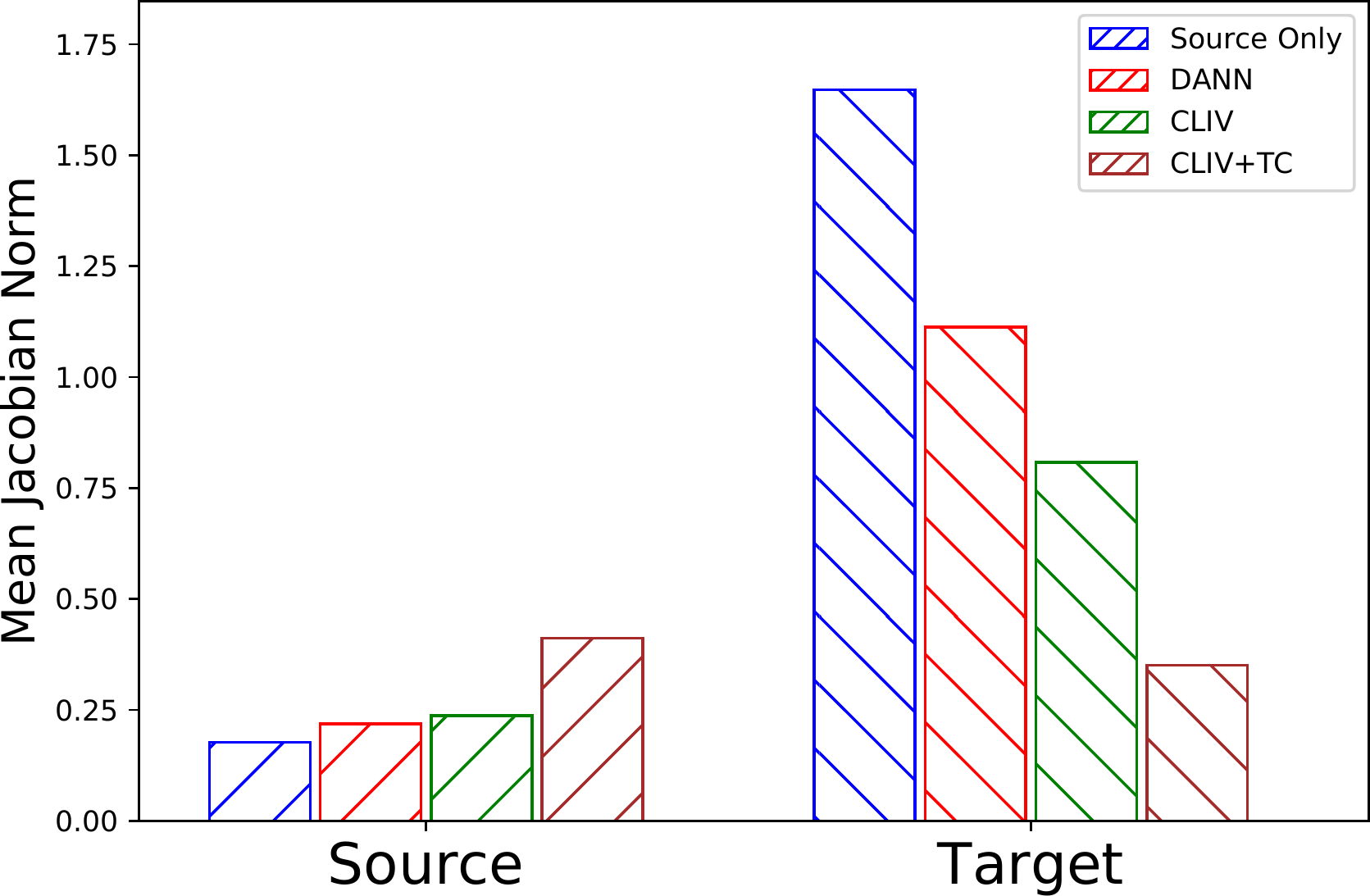}
  }
  \subfigure[\textbf{A $\shortrightarrow$ D}]{
    \label{fig:sensitivity_TC_AD}
    \includegraphics[width=0.46\textwidth]{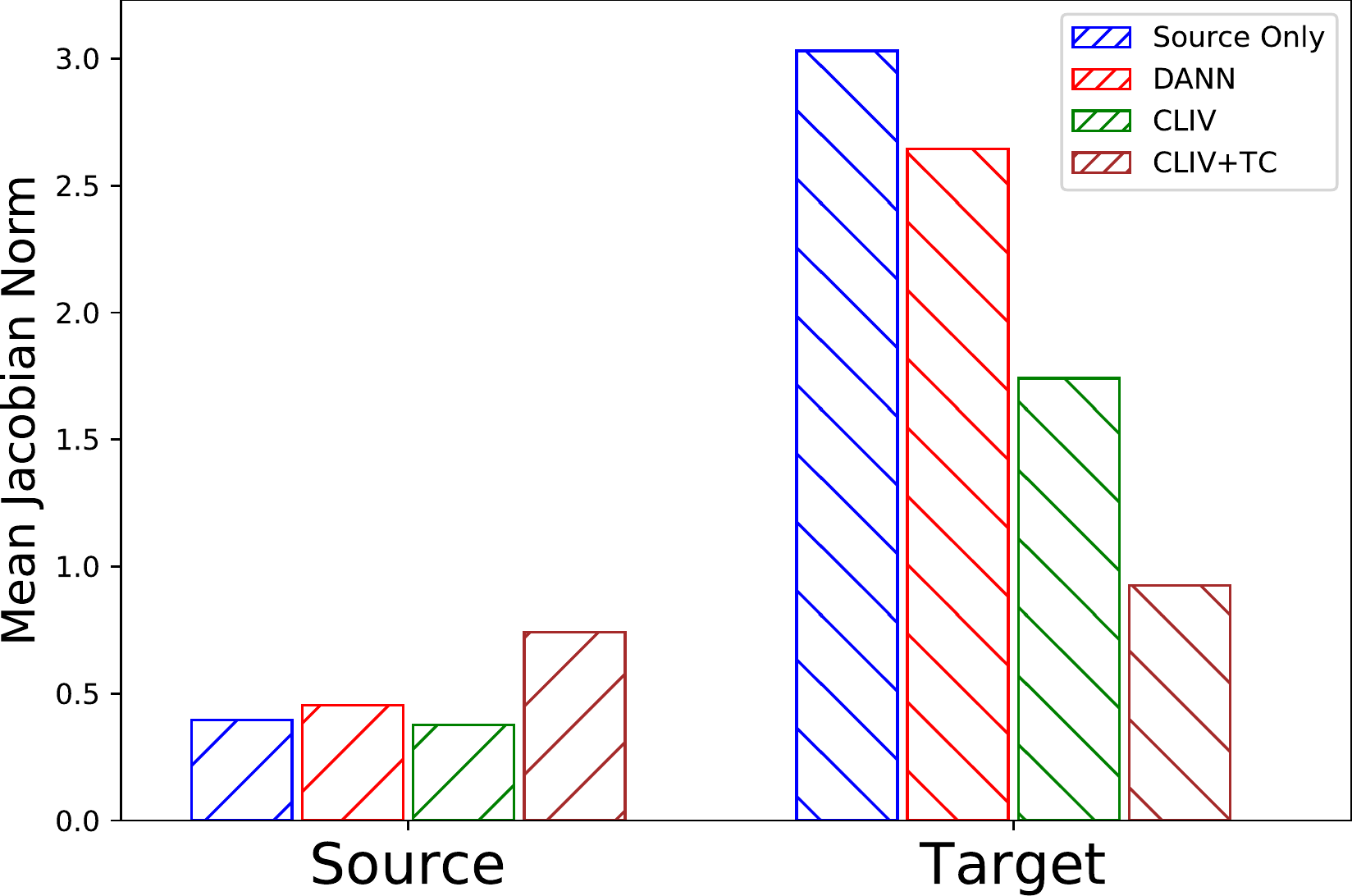}
  }
  \vspace{-0.1in}
  \caption{The effect of TC on the target and source sensitivity.}
  %\textbf{(a)} and \textbf{(b)} show the mean Jacobian norm on target and source domains with different methods.}
  \label{fig:sensitivity_TC}
\end{minipage}
\end{figure}

\textbf{Sensitivity Analysis.} To investigate the impact of TC on the model sensitivity, we compare the mean Jacobian norm of models trained with various objectives (\cref{fig:sensitivity_TC}). TC coupled with CLIV greatly improves the model's robustness on target, with a small increase of the source sensitivity. %since improving the joint classifier on target comes at the expense of its performance on source.

%\paragraph{Spectral Analysis.} To investigate the discriminability of the learned features, we apply singular value decomposition (SVD) to compute respectively all singular values and eigenvectors of the source and target feature matrices, in order to examine the effect of target consistency on the informative signals of eigenvectors corresponding to smaller singular values

\textbf{Ideal Joint Hypothesis and Distributions Discrepancy.} We evaluate the performances of the ideal joint hypothesis, which can be found by training an MLP classifier on top of a frozen features extractor on source and target data with labels. \cref{fig:lambda_and_discrepancy}(b) provides empirical evidence that TC produces a better joint hypothesis $h^{\lambda}$, thus more transferable representations. Additionally, as proxy measure of domain discrepancy \cite{ben2010theory}, we compute the A-distance, defined as $d_{A}=2(1-2 \varepsilon)$, with $\varepsilon$ as the error rate of a domain classifier trained to discriminate source and target domains. \cref{fig:lambda_and_discrepancy}(a) shows that TC decreases $d_{A}$, implying a better invariance.

\begin{figure}[!htbp]
\vspace{-5pt}
\centering
\begin{minipage}{0.6\textwidth}
    \centering
      \includegraphics[width=0.95\textwidth]{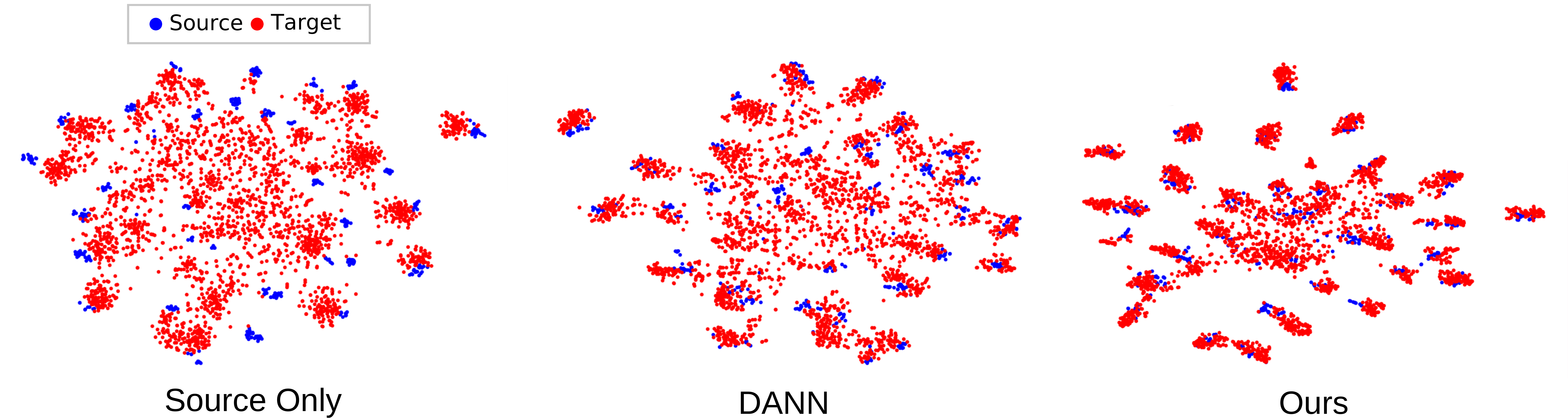}
      \caption{t-SNE of the adapted features (D $\shortrightarrow$ A) of, \textit{left}: ResNet-50, \textit{center}: DANN, \textit{right}: Ours.}
      \vspace{-0.15in}
      \label{fig:TSNE}
\end{minipage}%
\hfill
\begin{minipage}{0.37\textwidth}
    \centering
      \includegraphics[width=0.9\textwidth]{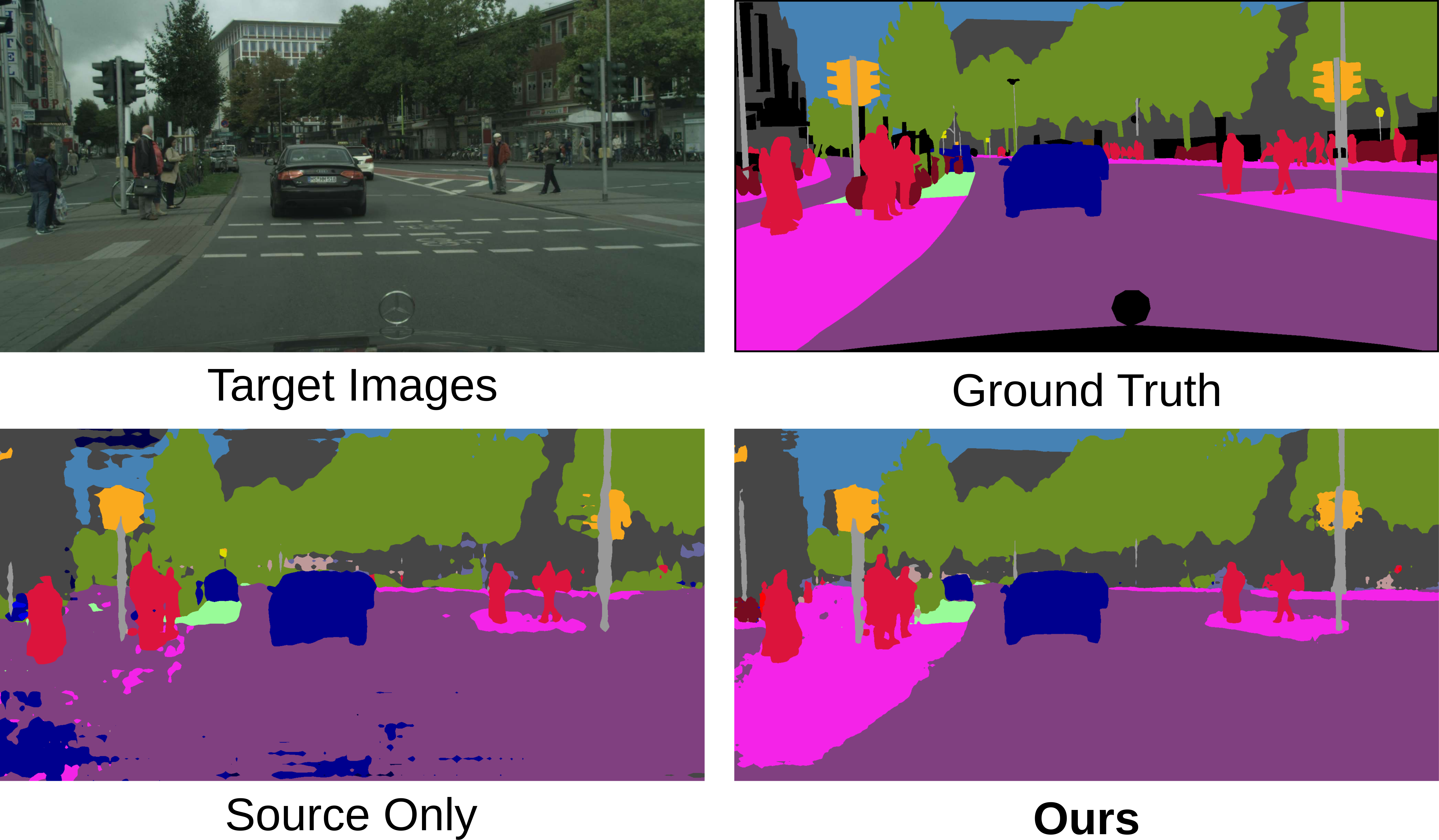}
      \vspace{-0.07in}
      \caption{Qualitative Results on GTA5 $\shortrightarrow$ Cityscapes.}
      \vspace{-0.15in}
      \label{fig:qualitative_segmentation}
\end{minipage}
\end{figure}

\textbf{Qualitative Analysis.}
We visualize the feature representations of \textbf{D $\shortrightarrow$ A}  task of Office-31 with t-SNE \cite{maaten2008visualizing} in \cref{fig:TSNE}. Our method produces a well aligned source and target features showing the benefits of coupling consistency and class level discrimination. The method also helps obtaining locally consistent and globally coherent predictions for semantic segmentation as illustrated in \cref{fig:qualitative_segmentation}.

\section{Related Work}

\textbf{Domain Adaptation.} The covariate shift adaptation has been studied by \cite{huang2007correcting,gretton2009covariate,sugiyama2008direct,sugiyama2007covariate} and label shift with kernel mean matching \cite{zhang2013domain,du2014semi} and Optimal Transport \cite{redko2018optimal}. Since Importance Sampling based methods are limited to distributions which share enough statistical support \cite{johansson2019support, d2017overlap}, an important line of works focuses on learning domain Invariant Representations (IR) \cite{ganin2015unsupervised,long2015learning} for reconciling two non-overlapping data distributions. IR has led to a furnished literature; \textit{Joint Adaptation Network} which aligns joint distribution of representations across layers \cite{long2016unsupervised}, \textit{Conditional Domain Adaptation Network} which performs the multilinear conditioning between representations and predictions \cite{long2018conditional}. Recently, significant progress have been made towards learning more transferable representations. In the work \cite{liu2019transferable}, it has been shown that invariance can be achieved by generating intermediate consistent representations, preserving their transferability. \cite{chen2019transferability} have brought to light that invariance often lead to poor discriminability of features, characterized by low rank representations. Therefore, they suggest to penalize the highest singular value of a batch of representations. \cite{wang2019transferable} have revisited the principle of batch normalization by building a transferable layer which aligns naturally mean and variance of representations across domains. 

%Domain adaptation (DA) \cite{pan2009survey}is a type of transductive transfer learning where the target task remains the same as the source, but the domain differs. The objective of DA is to train a learner to generalize across different domains of different distributions by reducing the shift across the source and the target domain. DA methods can be grouped into two classes, moment matching methods \cite{gretton2012optimal,long2015learning,long2017deep}, that aligns the representations by minimizing the distribution discrepancy in the features space, and adversarial methods, which introduces a domain discriminator to distinguish between source and target features, and push the feature extractor to produce similar representation across domains. DANN \cite{ganin2016domain} first proposed such a setting for DA, and later was extended to a conditional setting by CDAN \cite{long2018conditional} and with multiple discriminators for multimodal representation alignment by MADA \cite{pei2018multi}.

\textbf{Consistency Regularization.}
Consistency based semi-supervised methods \cite{laine2016temporal,tarvainen2017mean,berthelot2019mixmatch,verma2019interpolation,sohn2020fixmatch} have enjoyed great success in recent years, closing the gap with their fully supervised counterparts. Such methods are based on a simple concept: the prediction function should produce similar outputs for similar inputs, and by enforcing such a constraint, the resulting decision boundary will lie in low density regions echoing a more robust model. Semantically similar inputs can be obtained by a simple Gaussian noise injection \cite{laine2016temporal, tarvainen2017mean}, data augmentations \cite{berthelot2019mixmatch, sohn2020fixmatch}, or adversarial attacks \cite{miyato2018virtual}. The regularization term added to the training objective consists of a distance measure (\eg $L^2$,  KL divergence) between the function's output of a clean and a perturbed input.

To our knowledge, DIRT-T \cite{shu2018dirt} is the first work which incorporates consistency with invariant representations. Our work differs with it by exploring the connection between class-level invariance and target consistency, without the need to enforce a source consistency. Recently, \textit{Transferable Adversarial Training} \cite{liu2019transferable} explores the role of consistency in the representation space for bridging the distributional gap without hurting transferability. Our approach investigates consistency \textit{w.r.t} perturbations in the input space as a strong inductive bias for improving representations transferability. 
\vspace{-0.15in}

\section{Conclusion and Future Work}
%We have presented a new method for UDA which enforces the cluster assumption in the target domain when learning domain invariant representations. Our approach addresses the lack of robustness of domain adversarial learning by promoting consistent predictions, named \textit{Target Consistency} (TC),  to a various set of perturbations in the target domain. Crucially, our findings show that TC interacts strongly with Class-Level InVariance (CLIV) of representations and thus improves substantially their transferability. Through extensive experiments, we have shown that our approach outperforms significantly other methods based on invariant representations, validating our analysis. Furthermore, we have demonstrated the generality of TC by achieving competitive performances on a segmentation task. Finally, TC has the advantage to be orthogonal to recent works \cite{liu2019transferable, chen2019transferability} for improving transferability of invariant representations. Thus, combining them is an interesting research direction. In a  future work, we will explore how TC can mitigate the negative transfer in the challenging situation of label shift \cite{zhao2019learning}. Additionally, since our analysis is limited to perturbations in the image space, we plan to study language perturbations for NLP applications.

In this work, we presented a new approach to address
the lack of robustness of domain adversarial learning by promoting consistent predictions, named \textit{Target Consistency} (TC),  to a set of various input perturbations in the target domain. Crucially, we show that TC interacts strongly with Class-Level InVariance (CLIV) of representations, substantially improving their transferability. Through extensive experiments, our approach outperforms other methods based on invariant representations, validating our analysis. Finally, TC has the advantage of being orthogonal to recent works \cite{liu2019transferable, chen2019transferability} for improving transferability of invariant representations. Thus, combining them is an interesting research direction. In a future work, we will explore how TC can mitigate the negative transfer in the challenging situation of label shift \cite{zhao2019learning}. Additionally, since our analysis is limited to perturbations in the image space, we plan to study language perturbations for NLP applications.

\section*{Broader Impact Statement}
In industrial applications of Machine Learning, the setup is often restricted to very limited access of annotated data. Indeed, large-scale annotation is time-consuming, cost-prohibitive and difficult to assess its quality. On the one hand, Domain Adaptation is of great interest by allowing to adapt a model to new data distribution. It only asks labeled data from a related source distribution (\eg a controlled pipeline of data acquisition) and unlabeled data from the new distribution, which is often much cheaper to acquire.  In this context, the proposed method can be beneficial, based on its capability of leveraging large sets of annotated data to extract valuable training signal to better adapt the model.
On the other hand, the industrialization of adapted model may be hazardous since the absence of labeled target data does not allow to perform cross-validation (\ie no hyper-parameter selection or model selection) or to monitor model performances. Our method can alleviate such difficulty given its stability across a wide range of hyperparameter values, making it of particular interest for practitioners.

However, we point out the development of DA may expose society to new risks. The open access to large scale high quality annotated data, such as ImageNet, may offer valuable resources to ill-intention individuals. For instance, the \textbf{PersonX} dataset \cite{sun2019dissecting} is a dataset of synthetic 3D characters for which the task consists in estimating the viewpoint of the character on a scene. Coupling such synthetic data with large scale records of security cameras may lead to building more efficient individual tracking and be used for intrusive surveillance or military applications \eg drones. 

To conclude, the line of study of DA offers to bridge two datasets to address a specific task. Therefore, some data, which at first glance seems not exposed to a risk of wrong usage, may be leveraged, indirectly, for improving ill-intention ML applications.

\section*{Acknowledgements}
Victor Bouvier is funded by Sidetrade and ANRT (France) through a CIFRE
collaboration with CentraleSupélec. Yassine Ouali is supported by Randstad corporate research chair in collaboration with CentraleSupélec.
This work was performed using HPC resources from the Mésocentre computing center of
CentraleSupélec and École Normale Supérieure Paris-Saclay supported by CNRS and Région \^Ile-de-France (\url{http://mesocentre.centralesupelec.fr/}) and Saclay-IA plateform of Université Paris-Saclay.

%\href{https://neurips2019creativity.github.io/doc/MidiMe_%20Personalizing%20a%20MusicVAE%20model%20with%20user%20data.pdf}{1}
%\href{https://neurips2019creativity.github.io/doc/Text%20Conditional%20Lyric%20Video%20Generation.pdf}{2}

\begin{small}
\bibliographystyle{ieee}
\bibliography{bibliography}

\begin{thebibliography}{10}\itemsep=-1pt

\bibitem{amodei2016concrete}
D.~Amodei, C.~Olah, J.~Steinhardt, P.~Christiano, J.~Schulman, and D.~Man{\'e}.
\newblock Concrete problems in ai safety.
\newblock {\em arXiv preprint arXiv:1606.06565}, 2016.

\bibitem{arjovsky2019invariant}
M.~Arjovsky, L.~Bottou, I.~Gulrajani, and D.~Lopez-Paz.
\newblock Invariant risk minimization.
\newblock {\em arXiv preprint arXiv:1907.02893}, 2019.

\bibitem{beery2018recognition}
S.~Beery, G.~Van~Horn, and P.~Perona.
\newblock Recognition in terra incognita.
\newblock In {\em Proceedings of the European Conference on Computer Vision
  (ECCV)}, pages 456--473, 2018.

\bibitem{ben2010theory}
S.~Ben-David, J.~Blitzer, K.~Crammer, A.~Kulesza, F.~Pereira, and J.~W.
  Vaughan.
\newblock A theory of learning from different domains.
\newblock {\em Machine learning}, 79(1-2):151--175, 2010.

\bibitem{ben2007analysis}
S.~Ben-David, J.~Blitzer, K.~Crammer, and F.~Pereira.
\newblock Analysis of representations for domain adaptation.
\newblock In {\em Advances in neural information processing systems}, pages
  137--144, 2007.

\bibitem{berthelot2019remixmatch}
D.~Berthelot, N.~Carlini, E.~D. Cubuk, A.~Kurakin, K.~Sohn, H.~Zhang, and
  C.~Raffel.
\newblock Remixmatch: Semi-supervised learning with distribution matching and
  augmentation anchoring.
\newblock In {\em 8th International Conference on Learning Representations,
  {ICLR} 2020, Addis Ababa, Ethiopia, April 26-30, 2020}. OpenReview.net, 2020.

\bibitem{berthelot2019mixmatch}
D.~Berthelot, N.~Carlini, I.~Goodfellow, N.~Papernot, A.~Oliver, and C.~A.
  Raffel.
\newblock Mixmatch: A holistic approach to semi-supervised learning.
\newblock In {\em Advances in Neural Information Processing Systems}, pages
  5050--5060, 2019.

\bibitem{bouvier2020robust}
V.~Bouvier, P.~Very, C.~Chastagnol, M.~Tami, and C.~Hudelot.
\newblock Robust domain adaptation: Representations, weights and inductive
  bias.
\newblock {\em arXiv preprint arXiv:2006.13629}, 2020.

\bibitem{cao2018partial}
Z.~Cao, L.~Ma, M.~Long, and J.~Wang.
\newblock Partial adversarial domain adaptation.
\newblock In {\em Proceedings of the European Conference on Computer Vision
  (ECCV)}, pages 135--150, 2018.

\bibitem{carmon2019unlabeled}
Y.~Carmon, A.~Raghunathan, L.~Schmidt, J.~C. Duchi, and P.~S. Liang.
\newblock Unlabeled data improves adversarial robustness.
\newblock In {\em Advances in Neural Information Processing Systems}, pages
  11190--11201, 2019.

\bibitem{4787647}
O.~{Chapelle}, B.~{Scholkopf}, and A.~{Zien, Eds.}
\newblock Semi-supervised learning (chapelle, o. et al., eds.; 2006) [book
  reviews].
\newblock {\em IEEE Transactions on Neural Networks}, 20(3):542--542, 2009.

\bibitem{chen2017deeplab}
L.-C. Chen, G.~Papandreou, I.~Kokkinos, K.~Murphy, and A.~L. Yuille.
\newblock Deeplab: Semantic image segmentation with deep convolutional nets,
  atrous convolution, and fully connected crfs.
\newblock {\em IEEE transactions on pattern analysis and machine intelligence},
  40(4):834--848, 2017.

\bibitem{chen2019transferability}
X.~Chen, S.~Wang, M.~Long, and J.~Wang.
\newblock Transferability vs. discriminability: Batch spectral penalization for
  adversarial domain adaptation.
\newblock In {\em International Conference on Machine Learning}, pages
  1081--1090, 2019.

\bibitem{combes2020domain}
R.~T.~d. Combes, H.~Zhao, Y.-X. Wang, and G.~Gordon.
\newblock Domain adaptation with conditional distribution matching and
  generalized label shift.
\newblock {\em arXiv preprint arXiv:2003.04475}, 2020.

\bibitem{cordts2016cityscapes}
M.~Cordts, M.~Omran, S.~Ramos, T.~Rehfeld, M.~Enzweiler, R.~Benenson,
  U.~Franke, S.~Roth, and B.~Schiele.
\newblock The cityscapes dataset for semantic urban scene understanding.
\newblock In {\em Proceedings of the IEEE conference on computer vision and
  pattern recognition}, pages 3213--3223, 2016.

\bibitem{cubuk2019autoaugment}
E.~D. Cubuk, B.~Zoph, D.~Mane, V.~Vasudevan, and Q.~V. Le.
\newblock Autoaugment: Learning augmentation strategies from data.
\newblock In {\em Proceedings of the IEEE conference on computer vision and
  pattern recognition}, pages 113--123, 2019.

\bibitem{cubuk2019randaugment}
E.~D. Cubuk, B.~Zoph, J.~Shlens, and Q.~V. Le.
\newblock Randaugment: Practical data augmentation with no separate search.
\newblock {\em arXiv preprint arXiv:1909.13719}, 2019.

\bibitem{d2017overlap}
A.~D'Amour, P.~Ding, A.~Feller, L.~Lei, and J.~Sekhon.
\newblock Overlap in observational studies with high-dimensional covariates.
\newblock {\em arXiv preprint arXiv:1711.02582}, 2017.

\bibitem{deng2009imagenet}
J.~Deng, W.~Dong, R.~Socher, L.-J. Li, K.~Li, and L.~Fei-Fei.
\newblock Imagenet: A large-scale hierarchical image database.
\newblock In {\em 2009 IEEE conference on computer vision and pattern
  recognition}, pages 248--255. Ieee, 2009.

\bibitem{du2014semi}
M.~C. Du~Plessis and M.~Sugiyama.
\newblock Semi-supervised learning of class balance under class-prior change by
  distribution matching.
\newblock {\em Neural Networks}, 50:110--119, 2014.

\bibitem{ganin2015unsupervised}
Y.~Ganin and V.~Lempitsky.
\newblock Unsupervised domain adaptation by backpropagation.
\newblock In {\em International Conference on Machine Learning}, pages
  1180--1189, 2015.

\bibitem{ganin2016domain}
Y.~Ganin, E.~Ustinova, H.~Ajakan, P.~Germain, H.~Larochelle, F.~Laviolette,
  M.~Marchand, and V.~Lempitsky.
\newblock Domain-adversarial training of neural networks.
\newblock {\em The Journal of Machine Learning Research}, 17(1):2096--2030,
  2016.

\bibitem{geva2019we}
M.~Geva, Y.~Goldberg, and J.~Berant.
\newblock Are we modeling the task or the annotator? an investigation of
  annotator bias in natural language understanding datasets.
\newblock {\em arXiv preprint arXiv:1908.07898}, 2019.

\bibitem{gretton2012kernel}
A.~Gretton, K.~M. Borgwardt, M.~J. Rasch, B.~Sch{\"o}lkopf, and A.~Smola.
\newblock A kernel two-sample test.
\newblock {\em Journal of Machine Learning Research}, 13(Mar):723--773, 2012.

\bibitem{gretton2009covariate}
A.~Gretton, A.~Smola, J.~Huang, M.~Schmittfull, K.~Borgwardt, and
  B.~Sch{\"o}lkopf.
\newblock Covariate shift by kernel mean matching.
\newblock {\em Dataset shift in machine learning}, 3(4):5, 2009.

\bibitem{he2016deep}
K.~He, X.~Zhang, S.~Ren, and J.~Sun.
\newblock Deep residual learning for image recognition.
\newblock In {\em Proceedings of the IEEE conference on computer vision and
  pattern recognition}, pages 770--778, 2016.

\bibitem{hendrycks2019augmix}
D.~Hendrycks, N.~Mu, E.~D. Cubuk, B.~Zoph, J.~Gilmer, and B.~Lakshminarayanan.
\newblock Augmix: {A} simple data processing method to improve robustness and
  uncertainty.
\newblock In {\em 8th International Conference on Learning Representations,
  {ICLR} 2020, Addis Ababa, Ethiopia, April 26-30, 2020}. OpenReview.net, 2020.

\bibitem{huang2007correcting}
J.~Huang, A.~Gretton, K.~Borgwardt, B.~Sch{\"o}lkopf, and A.~J. Smola.
\newblock Correcting sample selection bias by unlabeled data.
\newblock In {\em Advances in neural information processing systems}, pages
  601--608, 2007.

\bibitem{johansson2019support}
F.~Johansson, D.~Sontag, and R.~Ranganath.
\newblock Support and invertibility in domain-invariant representations.
\newblock In {\em The 22nd International Conference on Artificial Intelligence
  and Statistics}, pages 527--536, 2019.

\bibitem{kannan2018adversarial}
H.~Kannan, A.~Kurakin, and I.~J. Goodfellow.
\newblock Adversarial logit pairing.
\newblock {\em CoRR}, abs/1803.06373, 2018.

\bibitem{kingma2014adam}
D.~P. Kingma and J.~Ba.
\newblock Adam: A method for stochastic optimization.
\newblock {\em arXiv preprint arXiv:1412.6980}, 2014.

\bibitem{laine2016temporal}
S.~Laine and T.~Aila.
\newblock Temporal ensembling for semi-supervised learning.
\newblock {\em arXiv preprint arXiv:1610.02242}, 2016.

\bibitem{lim2019fast}
S.~Lim, I.~Kim, T.~Kim, C.~Kim, and S.~Kim.
\newblock Fast autoaugment.
\newblock In {\em Advances in Neural Information Processing Systems}, pages
  6662--6672, 2019.

\bibitem{liu2019transferable}
H.~Liu, M.~Long, J.~Wang, and M.~Jordan.
\newblock Transferable adversarial training: A general approach to adapting
  deep classifiers.
\newblock In {\em International Conference on Machine Learning}, pages
  4013--4022, 2019.

\bibitem{long2015learning}
M.~Long, Y.~Cao, J.~Wang, and M.~I. Jordan.
\newblock Learning transferable features with deep adaptation networks.
\newblock In {\em Proceedings of the 32nd International Conference on
  International Conference on Machine Learning-Volume 37}, pages 97--105. JMLR.
  org, 2015.

\bibitem{long2018conditional}
M.~Long, Z.~Cao, J.~Wang, and M.~I. Jordan.
\newblock Conditional adversarial domain adaptation.
\newblock In {\em Advances in Neural Information Processing Systems}, pages
  1640--1650, 2018.

\bibitem{long2016unsupervised}
M.~Long, H.~Zhu, J.~Wang, and M.~I. Jordan.
\newblock Unsupervised domain adaptation with residual transfer networks.
\newblock In {\em Advances in Neural Information Processing Systems}, pages
  136--144, 2016.

\bibitem{long2017deep}
M.~Long, H.~Zhu, J.~Wang, and M.~I. Jordan.
\newblock Deep transfer learning with joint adaptation networks.
\newblock In {\em Proceedings of the 34th International Conference on Machine
  Learning-Volume 70}, pages 2208--2217. JMLR. org, 2017.

\bibitem{maaten2008visualizing}
L.~v.~d. Maaten and G.~Hinton.
\newblock Visualizing data using t-sne.
\newblock {\em Journal of machine learning research}, 9(Nov):2579--2605, 2008.

\bibitem{mansour2009domain}
Y.~Mansour, M.~Mohri, and A.~Rostamizadeh.
\newblock Domain adaptation: Learning bounds and algorithms.
\newblock In {\em 22nd Conference on Learning Theory, COLT 2009}, 2009.

\bibitem{marcus2020next}
G.~Marcus.
\newblock The next decade in ai: Four steps towards robust artificial
  intelligence.
\newblock {\em arXiv preprint arXiv:2002.06177}, 2020.

\bibitem{miyato2018virtual}
T.~Miyato, S.-i. Maeda, M.~Koyama, and S.~Ishii.
\newblock Virtual adversarial training: a regularization method for supervised
  and semi-supervised learning.
\newblock {\em IEEE transactions on pattern analysis and machine intelligence},
  41(8):1979--1993, 2018.

\bibitem{NovakBAPS18}
R.~Novak, Y.~Bahri, D.~A. Abolafia, J.~Pennington, and J.~Sohl{-}Dickstein.
\newblock Sensitivity and generalization in neural networks: an empirical
  study.
\newblock In {\em 6th International Conference on Learning Representations,
  {ICLR} 2018, Vancouver, BC, Canada, April 30 - May 3, 2018, Conference Track
  Proceedings}. OpenReview.net, 2018.

\bibitem{pan2009survey}
S.~J. Pan and Q.~Yang.
\newblock A survey on transfer learning.
\newblock {\em IEEE Transactions on knowledge and data engineering},
  22(10):1345--1359, 2009.

\bibitem{paszke2019pytorch}
A.~Paszke, S.~Gross, F.~Massa, A.~Lerer, J.~Bradbury, G.~Chanan, T.~Killeen,
  Z.~Lin, N.~Gimelshein, L.~Antiga, et~al.
\newblock Pytorch: An imperative style, high-performance deep learning library.
\newblock In {\em Advances in Neural Information Processing Systems}, pages
  8024--8035, 2019.

\bibitem{peng2017visda}
X.~Peng, B.~Usman, N.~Kaushik, J.~Hoffman, D.~Wang, and K.~Saenko.
\newblock Visda: The visual domain adaptation challenge.
\newblock {\em arXiv preprint arXiv:1710.06924}, 2017.

\bibitem{quionero2009dataset}
J.~Quionero-Candela, M.~Sugiyama, A.~Schwaighofer, and N.~D. Lawrence.
\newblock {\em Dataset shift in machine learning}.
\newblock The MIT Press, 2009.

\bibitem{radford2015unsupervised}
A.~Radford, L.~Metz, and S.~Chintala.
\newblock Unsupervised representation learning with deep convolutional
  generative adversarial networks.
\newblock {\em arXiv preprint arXiv:1511.06434}, 2015.

\bibitem{redko2018optimal}
I.~Redko, N.~Courty, R.~Flamary, and D.~Tuia.
\newblock Optimal transport for multi-source domain adaptation under target
  shift.
\newblock {\em arXiv preprint arXiv:1803.04899}, 2018.

\bibitem{richter2016playing}
S.~R. Richter, V.~Vineet, S.~Roth, and V.~Koltun.
\newblock Playing for data: Ground truth from computer games.
\newblock In {\em European conference on computer vision}, pages 102--118.
  Springer, 2016.

\bibitem{saenko2010adapting}
K.~Saenko, B.~Kulis, M.~Fritz, and T.~Darrell.
\newblock Adapting visual category models to new domains.
\newblock In {\em European conference on computer vision}, pages 213--226.
  Springer, 2010.

\bibitem{sankaranarayanan2018generate}
S.~Sankaranarayanan, Y.~Balaji, C.~D. Castillo, and R.~Chellappa.
\newblock Generate to adapt: Aligning domains using generative adversarial
  networks.
\newblock In {\em Proceedings of the IEEE Conference on Computer Vision and
  Pattern Recognition}, pages 8503--8512, 2018.

\bibitem{shu2018dirt}
R.~Shu, H.~H. Bui, H.~Narui, and S.~Ermon.
\newblock A {DIRT-T} approach to unsupervised domain adaptation.
\newblock In {\em 6th International Conference on Learning Representations,
  {ICLR} 2018, Vancouver, BC, Canada, April 30 - May 3, 2018, Conference Track
  Proceedings}. OpenReview.net, 2018.

\bibitem{sohn2020fixmatch}
K.~Sohn, D.~Berthelot, C.-L. Li, Z.~Zhang, N.~Carlini, E.~D. Cubuk, A.~Kurakin,
  H.~Zhang, and C.~Raffel.
\newblock Fixmatch: Simplifying semi-supervised learning with consistency and
  confidence.
\newblock {\em arXiv preprint arXiv:2001.07685}, 2020.

\bibitem{sugiyama2007covariate}
M.~Sugiyama, M.~Krauledat, and K.-R. M{\~A}{\v{z}}ller.
\newblock Covariate shift adaptation by importance weighted cross validation.
\newblock {\em Journal of Machine Learning Research}, 8(May):985--1005, 2007.

\bibitem{sugiyama2008direct}
M.~Sugiyama, S.~Nakajima, H.~Kashima, P.~V. Buenau, and M.~Kawanabe.
\newblock Direct importance estimation with model selection and its application
  to covariate shift adaptation.
\newblock In {\em Advances in neural information processing systems}, pages
  1433--1440, 2008.

\bibitem{sun2019dissecting}
X.~Sun and L.~Zheng.
\newblock Dissecting person re-identification from the viewpoint of viewpoint.
\newblock In {\em Proceedings of the IEEE Conference on Computer Vision and
  Pattern Recognition}, pages 608--617, 2019.

\bibitem{tarvainen2017mean}
A.~Tarvainen and H.~Valpola.
\newblock Mean teachers are better role models: Weight-averaged consistency
  targets improve semi-supervised deep learning results.
\newblock In {\em Advances in neural information processing systems}, pages
  1195--1204, 2017.

\bibitem{torrey2010transfer}
L.~Torrey and J.~Shavlik.
\newblock Transfer learning.
\newblock In {\em Handbook of research on machine learning applications and
  trends: algorithms, methods, and techniques}, pages 242--264. IGI Global,
  2010.

\bibitem{tsai2018learning}
Y.-H. Tsai, W.-C. Hung, S.~Schulter, K.~Sohn, M.-H. Yang, and M.~Chandraker.
\newblock Learning to adapt structured output space for semantic segmentation.
\newblock In {\em Proceedings of the IEEE Conference on Computer Vision and
  Pattern Recognition}, pages 7472--7481, 2018.

\bibitem{tzeng2017adversarial}
E.~Tzeng, J.~Hoffman, K.~Saenko, and T.~Darrell.
\newblock Adversarial discriminative domain adaptation.
\newblock In {\em Proceedings of the IEEE Conference on Computer Vision and
  Pattern Recognition}, pages 7167--7176, 2017.

\bibitem{venkateswara2017deep}
H.~Venkateswara, J.~Eusebio, S.~Chakraborty, and S.~Panchanathan.
\newblock Deep hashing network for unsupervised domain adaptation.
\newblock In {\em Proceedings of the IEEE Conference on Computer Vision and
  Pattern Recognition}, pages 5018--5027, 2017.

\bibitem{verma2019interpolation}
V.~Verma, A.~Lamb, J.~Kannala, Y.~Bengio, and D.~Lopez-Paz.
\newblock Interpolation consistency training for semi-supervised learning.
\newblock {\em arXiv preprint arXiv:1903.03825}, 2019.

\bibitem{vu2019advent}
T.-H. Vu, H.~Jain, M.~Bucher, M.~Cord, and P.~P{\'e}rez.
\newblock Advent: Adversarial entropy minimization for domain adaptation in
  semantic segmentation.
\newblock In {\em Proceedings of the IEEE Conference on Computer Vision and
  Pattern Recognition}, pages 2517--2526, 2019.

\bibitem{wang2019transferable}
X.~Wang, Y.~Jin, M.~Long, J.~Wang, and M.~I. Jordan.
\newblock Transferable normalization: Towards improving transferability of deep
  neural networks.
\newblock In {\em Advances in Neural Information Processing Systems}, pages
  1951--1961, 2019.

\bibitem{wu2019domain}
Y.~Wu, E.~Winston, D.~Kaushik, and Z.~Lipton.
\newblock Domain adaptation with asymmetrically-relaxed distribution alignment.
\newblock In {\em International Conference on Machine Learning}, pages
  6872--6881, 2019.

\bibitem{xie2019unsupervised}
Q.~Xie, Z.~Dai, E.~Hovy, M.-T. Luong, and Q.~V. Le.
\newblock Unsupervised data augmentation.
\newblock {\em arXiv preprint arXiv:1904.12848}, 2019.

\bibitem{yin2019fourier}
D.~Yin, R.~G. Lopes, J.~Shlens, E.~D. Cubuk, and J.~Gilmer.
\newblock A fourier perspective on model robustness in computer vision.
\newblock In {\em Advances in Neural Information Processing Systems}, pages
  13255--13265, 2019.

\bibitem{you2019universal}
K.~You, M.~Long, Z.~Cao, J.~Wang, and M.~I. Jordan.
\newblock Universal domain adaptation.
\newblock In {\em Proceedings of the IEEE Conference on Computer Vision and
  Pattern Recognition}, pages 2720--2729, 2019.

\bibitem{zhang2013domain}
K.~Zhang, B.~Sch{\"o}lkopf, K.~Muandet, and Z.~Wang.
\newblock Domain adaptation under target and conditional shift.
\newblock In {\em International Conference on Machine Learning}, pages
  819--827, 2013.

\bibitem{zhao2019learning}
H.~Zhao, R.~T. Des~Combes, K.~Zhang, and G.~Gordon.
\newblock On learning invariant representations for domain adaptation.
\newblock In {\em International Conference on Machine Learning}, pages
  7523--7532, 2019.

\bibitem{zheng2016improving}
S.~Zheng, Y.~Song, T.~Leung, and I.~Goodfellow.
\newblock Improving the robustness of deep neural networks via stability
  training.
\newblock In {\em Proceedings of the ieee conference on computer vision and
  pattern recognition}, pages 4480--4488, 2016.

\end{thebibliography}
\end{small}

\clearpage

\renewcommand{\thesubsection}{\Alph{subsection}}

\section*{\centering Supplementary Material}

\subsection{Detailed Results}
\label{sec:results}

\begin{table*}[htbp]
  \addtolength{\tabcolsep}{2pt}
  \centering
  \caption{Accuracy (\%) on {Office-31} for unsupervised domain adaptation (ResNet-50)}
  \label{table:todo}
  \resizebox{\textwidth}{!}{%
  \begin{tabular}{lccccccc}
    \toprule
    Method & A $\rightarrow$ W & D $\rightarrow$ W & W $\rightarrow$ D & A $\rightarrow$ D & D $\rightarrow$ A & W $\rightarrow$ A & Avg \\
    \midrule
   	ResNet-50 \cite{he2016deep} & 68.4$\pm$0.2 & 96.7$\pm$0.1 & 99.3$\pm$0.1 & 68.9$\pm$0.2 & 62.5$\pm$0.3 & 60.7$\pm$0.3 & 76.1 \\
    DAN \cite{long2015learning} & 80.5$\pm$0.4 & 97.1$\pm$0.2 & 99.6$\pm$0.1 & 78.6$\pm$0.2 & 63.6$\pm$0.3 & 62.8$\pm$0.2 & 80.4 \\
    RTN \cite{long2016unsupervised} & 84.5$\pm$0.2 & 96.8$\pm$0.1 & 99.4$\pm$0.1 & 77.5$\pm$0.3 & 66.2$\pm$0.2 & 64.8$\pm$0.3 & 81.6 \\
    DANN \cite{ganin2016domain} & 82.0$\pm$0.4 & 96.9$\pm$0.2 & 99.1$\pm$0.1 & 79.7$\pm$0.4 & 68.2$\pm$0.4 & 67.4$\pm$0.5 & 82.2 \\
    ADDA \cite{tzeng2017adversarial} & 86.2$\pm$0.5 & 96.2$\pm$0.3 & 98.4$\pm$0.3 & 77.8$\pm$0.3 & 69.5$\pm$0.4 & 68.9$\pm$0.5 & 82.9 \\
    JAN \cite{long2017deep} & 85.4$\pm$0.3 & {97.4}$\pm$0.2 & {99.8}$\pm$0.2 & 84.7$\pm$0.3 & 68.6$\pm$0.3 & 70.0$\pm$0.4 & 84.3 \\
    GTA \cite{sankaranarayanan2018generate} & 89.5$\pm$0.5 & 97.9$\pm$0.3 & 99.8$\pm$0.4 & 87.7$\pm$0.5 & 72.8$\pm$0.3 & 71.4$\pm$0.4 & 86.5 \\
    CDAN \cite{long2018conditional} & 94.1$\pm$0.1 & 98.6$\pm$0.1 & \textbf{100.0$\pm$.0} & 92.9$\pm$0.2 & 71.0$\pm$0.3 & 69.3$\pm$0.3 & 87.7 \\
    TAT \cite{liu2019transferable} & 92.5$\pm$0.3 & 99.3$\pm$0.1 & \textbf{100.0$\pm$.0} & 93.2$\pm$0.2 & 73.1$\pm$0.3 & 72.1$\pm$0.3 & 88.4 \\
    BSP \cite{chen2019transferability} & 93.3$\pm$0.2 & 98.2$\pm$0.2 & \textbf{100.0$\pm$.0} & 93.0$\pm$0.2 & 73.6$\pm$0.3 & 72.6$\pm$0.3 & 88.5 \\
    TransNorm \cite{wang2019transferable} & 95.7$\pm$0.5 & 98.7$\pm$0.3 & \textbf{100.0$\pm$.0} & 94.0$\pm$0.2 & 73.4$\pm$0.4 & 74.2$\pm$0.3 & 89.3 \\
    \textbf{Ours} & 94.8$\pm0.8$ & \textbf{99.1$\pm0.2$} & \textbf{100.0$\pm$.0} & 93.6$\pm0.9$ & \textbf{76.8$\pm1.3$} & 73.4$\pm0.7$ & \textbf{89.6} \\    
    \bottomrule
  \end{tabular}
  }
\end{table*}

\begin{table*}[!ht]
  \addtolength{\tabcolsep}{3pt}
  \centering
  \caption{Accuracy (\%) on {ImageCLEF-DA} for unsupervised domain adaptation (ResNet-50)}
  \label{table:todo}
  \resizebox{\textwidth}{!}{%
  \begin{tabular}{lccccccc}
    \toprule
    Method & I $\rightarrow$ P & P $\rightarrow$ I & I $\rightarrow$ C & C $\rightarrow$ I & C $\rightarrow$ P & P $\rightarrow$ C & Avg \\
    \midrule
    ResNet-50 \cite{he2016deep} & 74.8$\pm$0.3 & 83.9$\pm$0.1 & 91.5$\pm$0.3 & 78.0$\pm$0.2 & 65.5$\pm$0.3 & 91.2$\pm$0.3 & 80.7 \\
    DAN \cite{long2015learning} & 74.5$\pm$0.4 & 82.2$\pm$0.2 & 92.8$\pm$0.2 & 86.3$\pm$0.4 & 69.2$\pm$0.4 & 89.8$\pm$0.4 & 82.5 \\
    DANN \cite{ganin2016domain} & 75.0$\pm$0.6 & 86.0$\pm$0.3 & 96.2$\pm$0.4 & 87.0$\pm$0.5 & 74.3$\pm$0.5 & 91.5$\pm$0.6 & 85.0 \\
    JAN \cite{long2017deep} & {76.8}$\pm$0.4 & {88.0}$\pm$0.2 & {94.7}$\pm$0.2 & {89.5}$\pm$0.3 & {74.2}$\pm$0.3 & {91.7}$\pm$0.3 & 85.8 \\
    CDAN \cite{long2018conditional} & 77.7$\pm$0.3 & 90.7$\pm$0.2 & 97.7$\pm$0.3 & 91.3$\pm$0.3 & 74.2$\pm$0.2 & 94.3$\pm$0.3 & 87.7 \\
    TransNorm \cite{wang2019transferable} & 78.3$\pm$0.3 & 90.8$\pm$0.2 & 96.7$\pm$0.4 & 92.3$\pm$0.2 & 78.0$\pm$0.1 & 94.8$\pm$0.3 & 88.5 \\
    TAT \cite{liu2019transferable} & 78.8$\pm$0.2 & 92.0$\pm$0.2 & 97.5$\pm$0.3 & 92.0$\pm$0.3 & 78.2$\pm$0.4 & 94.7$\pm$0.4 & 88.9 \\
    \textbf{Ours} & \textbf{79.5$\pm0.4$} & \textbf{92.7$\pm0.3$} & \textbf{97.6$\pm0.2$} & \textbf{93.2$\pm0.4$} & \textbf{78.6$\pm0.2$} & \textbf{95.5$\pm0.4$} &  \textbf{89.5}
 \\
    \bottomrule
  \end{tabular}
  }
\end{table*}

\null

\begin{table*}[!ht]
  \addtolength{\tabcolsep}{-5pt}
  \centering
  \caption{Accuracy (\%) on {Office-Home} for unsupervised domain adaptation (ResNet-50)}
  \label{table:todo}
  \resizebox{\textwidth}{!}{%
  \begin{tabular}{lccccccccccccc}
    \toprule
    Method & Ar$\shortrightarrow$Cl & Ar$\shortrightarrow$Pr & Ar$\shortrightarrow$Rw & Cl$\shortrightarrow$Ar & Cl$\shortrightarrow$Pr & Cl$\shortrightarrow$Rw & Pr$\shortrightarrow$Ar & Pr$\shortrightarrow$Cl & Pr$\shortrightarrow$Rw & Rw$\shortrightarrow$Ar & Rw$\shortrightarrow$Cl & Rw$\shortrightarrow$Pr & Avg \\
    \midrule
    ResNet-50 \cite{he2016deep} & 34.9 & 50.0 & 58.0 & 37.4 & 41.9 & 46.2 & 38.5 & 31.2 & 60.4 & 53.9 & 41.2 & 59.9 & 46.1 \\
    DAN \cite{long2015learning} & 43.6 & 57.0 & 67.9 & 45.8 & 56.5 & 60.4 & 44.0 & 43.6 & 67.7 & 63.1 & 51.5 & 74.3 & 56.3 \\
    DANN \cite{ganin2016domain} & 45.6 & 59.3 & 70.1 & 47.0 & 58.5 & 60.9 & 46.1 & 43.7 & 68.5 & 63.2 & 51.8 & 76.8 & 57.6 \\
    JAN \cite{long2017deep} & 45.9 & 61.2 & 68.9 & 50.4 & 59.7 & 61.0 & 45.8 & 43.4 & 70.3 & 63.9 & 52.4 & 76.8 & 58.3 \\
    CDAN \cite{long2018conditional} & 50.7 & 70.6 & 76.0 & 57.6 & 70.0 & 70.0 & 57.4 & 50.9 & 77.3 & 70.9 & 56.7 & 81.6 & 65.8 \\
    TAT \cite{liu2019transferable} & 51.6 & 69.5 & 75.4 & 59.4 & 69.5 & 68.6 & 59.5 & 50.5 & 76.8 & 70.9 & 56.6 & 81.6 & 65.8 \\
	BSP \cite{chen2019transferability} & 52.0 & 68.6 & 76.1 & 58.0 & 70.3 & 70.2 & 58.6 & 50.2 & 77.6 & 72.2 & 59.3 & 81.9 & 66.3 \\
	TransNorm \cite{wang2019transferable} & 50.2 & 71.4 & 77.4 & 59.3 & \textbf{72.7} & \textbf{73.1} & 61.0 & 53.1 & 79.5 & 71.9 & 59.0 & 82.9 & 67.6 \\
    \textbf{Ours} & \textbf{53.1} & \textbf{73.0} & \textbf{77.0} & \textbf{62.6} & 72.4 & \textbf{73.1} & \textbf{63.8} & \textbf{54.4} & \textbf{79.8} & \textbf{74.6} & \textbf{60.4} & \textbf{83.3} & \textbf{69.0} \\    
    \bottomrule
  \end{tabular}%
  }
\end{table*}

\null

\begin{table}[!ht]
  \addtolength{\tabcolsep}{6pt}
  \centering 
  \caption{Accuracy (\%) on VisDA-2017}
  \label{table:todo}
  \vspace{-5pt}
  \resizebox{0.8\textwidth}{!}{%
  \begin{tabular}{lc|lc}
  	\toprule
  	\multicolumn{2}{c|}{ResNet-50} & \multicolumn{2}{c}{ResNet-101}\\
    Method & Synthetic $\rightarrow$ Real & Method & Synthetic $\rightarrow$ Real \\
    \midrule
    JAN \cite{long2017deep} & 61.6 & 					ResNet-101 \cite{he2016deep} & 52.4 \\
    GTA \cite{sankaranarayanan2018generate} & 69.5 & 	DANN \cite{ganin2016domain} & 57.4 \\
    CDAN \cite{long2018conditional} & 70.0 & 			CDAN \cite{long2018conditional} & 73.7 \\
    TAT \cite{liu2019transferable} & 71.9 & 			BSP \cite{chen2019transferability} & 75.9 \\
	\textbf{Ours}  & \textbf{77.5$\pm$0.7} &            \textbf{Ours} & \textbf{79.0$\pm$0.1} \\
    \bottomrule
  \end{tabular}
  }
\end{table}

\null

\begin{table*}[!ht]
  \centering
  \caption{Accuracy (\%) on the 5 hardest {Office-Home} task for Target Consistency ablation (ResNet-50)}
  \label{table:todo}
  \resizebox{1\textwidth}{!}{%
  \begin{tabular}{lcccccc}
    \toprule
    & Ar$\shortrightarrow$Cl & Cl$\shortrightarrow$Ar & Pr$\shortrightarrow$Ar & Pr$\shortrightarrow$Cl & Rw$\shortrightarrow$Cl & Avg \\
    \midrule
    $\mathcal L_{\mathrm{DANN}}$ & 45.2$\pm$0.7 & 48.8$\pm$0.5 & 46.8$\pm$0.2 & 43.5$\pm$0.3 & 53.6$\pm$0.3 &  47.6 \\ 
    $\mathcal L_{\mathrm{DANN}} + \mathcal L_{\mathrm{VAT}}$ & 44.3$\pm$0.2 & 50.3$\pm$1.8 & 48.5$\pm$1.1 & 43.6$\pm$0.6 & 53.5$\pm$0.2 &  48.0 \\ 
    $\mathcal L_{\mathrm{DANN}} + \mathcal L_{\mathrm{AUG}}$ & 46.2$\pm$0.4 & 55.3$\pm$0.5 & 53.2$\pm$1.4 & 46.0$\pm$0.4 & 55.6$\pm$0.5 &  51.3 \\ 
    $\mathcal L_{\mathrm{DANN}} + \mathcal L_{\mathrm{VAT}}+ \mathcal L_{\mathrm{AUG}}$ & 46.3$\pm$0.6 & 53.5$\pm$1.0 & 54.7$\pm$0.7 & 46.2$\pm$0.7 & 56.3$\pm$0.9 &  51.4 \\ 
    $\mathcal L_{\mathrm{DANN}} + \mathcal L_{\mathrm{VAT}}+ \mathcal L_{\mathrm{AUG}}$ /w MT & 46.6$\pm$0.3 & 53.3$\pm$0.7 & 52.8$\pm$0.3 & 46.9$\pm$0.8 & 55.6$\pm$0.5 &  51.0 \\
    \bottomrule

    $\mathcal L_{\mathrm{CDAN}}$ & 50.3$\pm$0.1 & 54.6$\pm$0.7 & 55.8$\pm$0.6 & 49.3$\pm$0.2 & 56.9$\pm$0.1 &  53.4 \\ 
    $\mathcal L_{\mathrm{CDAN}} + \mathcal L_{\mathrm{VAT}}$ & 50.1$\pm$0.5 & 58.5$\pm$0.6 & 59.1$\pm$0.6 & 49.8$\pm$0.2 & 57.9$\pm$0.1 &  55.1 \\ 
    $\mathcal L_{\mathrm{CDAN}} + \mathcal L_{\mathrm{AUG}}$ & 51.0$\pm$0.2 & 57.3$\pm$0.5 & 61.0$\pm$0.7 & 50.8$\pm$0.2 & 58.4$\pm$0.5 &  55.7 \\ 
    $\mathcal L_{\mathrm{CDAN}} + \mathcal L_{\mathrm{VAT}}+ \mathcal L_{\mathrm{AUG}}$ & 51.5$\pm$0.2 & 60.9$\pm$0.3 & 61.4$\pm$0.9 & 51.7$\pm$0.2 & 59.1$\pm$0.5 &  56.9 \\ 
    $\mathcal L_{\mathrm{CDAN}} + \mathcal L_{\mathrm{VAT}}+ \mathcal L_{\mathrm{AUG}}$ /w MT & 51.3$\pm$0.9 & 59.0$\pm$0.4 & 60.0$\pm$0.5 & 51.8$\pm$0.2 & 57.9$\pm$0.3 & 56.0 \\
    \bottomrule
    
    $\mathcal L_{\mathrm{CLIV}}$ & 52.6$\pm${0.8} & 60.1$\pm${0.3} & 60.6$\pm${0.9} & 52.1$\pm${0.7} & 58.3$\pm${0.4} &  56.7 \\ 
    $\mathcal L_{\mathrm{CLIV}} + \mathcal L_{\mathrm{VAT}}$ & 52.4$\pm${0.6} & 60.1$\pm${0.5} & 61.2$\pm${0.9} & 53.1$\pm${0.2} & 58.9$\pm${0.8} &  57.1 \\ 
    $\mathcal L_{\mathrm{CLIV}} + \mathcal L_{\mathrm{AUG}}$ & 53.1$\pm${0.5} & 62.3$\pm${0.6} & 62.6$\pm${0.8} & 53.1$\pm${1.0} & 59.5$\pm${0.3} & 58.1 \\ 
    $\mathcal L_{\mathrm{CLIV}} + \mathcal L_{\mathrm{VAT}}+ \mathcal L_{\mathrm{AUG}}$ & 53.0$\pm${0.1} & \textbf{62.8$\pm${0.7}} & 62.8$\pm${0.2} & 53.8$\pm${0.8} & \textbf{60.8$\pm${0.8}} &  58.6 \\ 
    $\mathcal L_{\mathrm{CLIV}} + \mathcal L_{\mathrm{VAT}}+ \mathcal L_{\mathrm{AUG}}$ /w MT & \textbf{53.1$\pm${1.5}} & 62.6$\pm${0.1} & \textbf{63.8$\pm${0.7}} & \textbf{54.4$\pm${0.6}} & 60.4$\pm${0.6} &  \textbf{58.9} \\ 
    \bottomrule
  \end{tabular}%
  }
\end{table*}

\null

\begin{table*}[!ht]
\addtolength{\tabcolsep}{6pt}
\centering
  \caption{mIoU on \textbf{GTA5 $\shortrightarrow$ Cityscapes}. 
  {\footnotesize{\textit{AdvEnt+MinEnt}* is an ensemble of two models.}}}
\label{table:GTAfull}
\small
\renewcommand{\arraystretch}{1.1}
\setlength{\tabcolsep}{2.2pt}
\resizebox{1\textwidth}{!}{%
\begin{tabular}{lcccccccccccccccccccc}
	\toprule
	Method & \rotatebox{90}{road} & \rotatebox{90}{sidewalk} & \rotatebox{90}{building} & \rotatebox{90}{wall} & \rotatebox{90}{fence} & \rotatebox{90}{pole} & \rotatebox{90}{light} & \rotatebox{90}{sign} & \rotatebox{90}{veg} & \rotatebox{90}{terrain} & \rotatebox{90}{sky} & \rotatebox{90}{person} & \rotatebox{90}{rider} & \rotatebox{90}{car} & \rotatebox{90}{truck} & \rotatebox{90}{bus} & \rotatebox{90}{train} & \rotatebox{90}{mbike} & \rotatebox{90}{bike} & mIoU\\
	\midrule
	ResNet-101 \cite{he2016deep} & 75.8 & 16.8 & 77.2 & 12.5 & 21.0 & 25.5 & 30.1 & 20.1 & 81.3 & 24.6 & 70.3 & 53.8 & 26.4 & 49.9 & 17.2 & 25.9 & 6.5 & 25.3 & 36.0 & 36.6 \\
	Adapt-SegMap \cite{tsai2018learning} & 86.5 & 36.0 & 79.9 & 23.4 & 23.3 & 23.9 & 35.2 & 14.8 & 83.4 & 33.3 & 75.6 & 58.5 & 27.6 & 73.7 & 32.5 & 35.4 & 3.9 & 30.1 & 28.1 & 42.4 \\
	AdvEnt \cite{vu2019advent} & 89.9 & 36.5 & 81.6 & 29.2 & 25.2 & 28.5 & 32.3 & 22.4 & 83.9 & 34.0 & 77.1 & 57.4 & 27.9 & 83.7 & 29.4 & 39.1 & 1.5 & 28.4 & 23.3 & 43.8 \\
    \textbf{Ours} & \textbf{91.0} & \textbf{41.9} & \textbf{81.6} & \textbf{30.1} & 22.6 & 26.0 & 28.8 & 13.6 & 82.6 & \textbf{37.2} & \textbf{81.9} & 56.1 & \textbf{29.3} & \textbf{84.8} & \textbf{34.1} & \textbf{48.8} & 0.0 & 26.8 & \textbf{35.7} & 44.9 \\
    \midrule
	AdvEnt+MinEnt* \cite{vu2019advent} &89.4&33.1&{81.0}&{26.6}&\textbf{26.8}&\textbf{27.2}&\textbf{33.5}&\textbf{24.7}&\textbf{83.9}&\textbf{36.7}&{78.8}&\textbf{58.7}&\textbf{30.5}&\textbf{84.8}&\textbf{38.5}&\textbf{44.5}&\textbf{1.7}&\textbf{31.6}&32.4&\textbf{45.5}\\
	\bottomrule
\end{tabular}
}
\end{table*}

\subsection{Experimental Details}
\label{sec:exp_details}

\subsubsection{Augmentations}

\begin{figure}[!ht]
  \centering
  \includegraphics[width=0.8\textwidth]{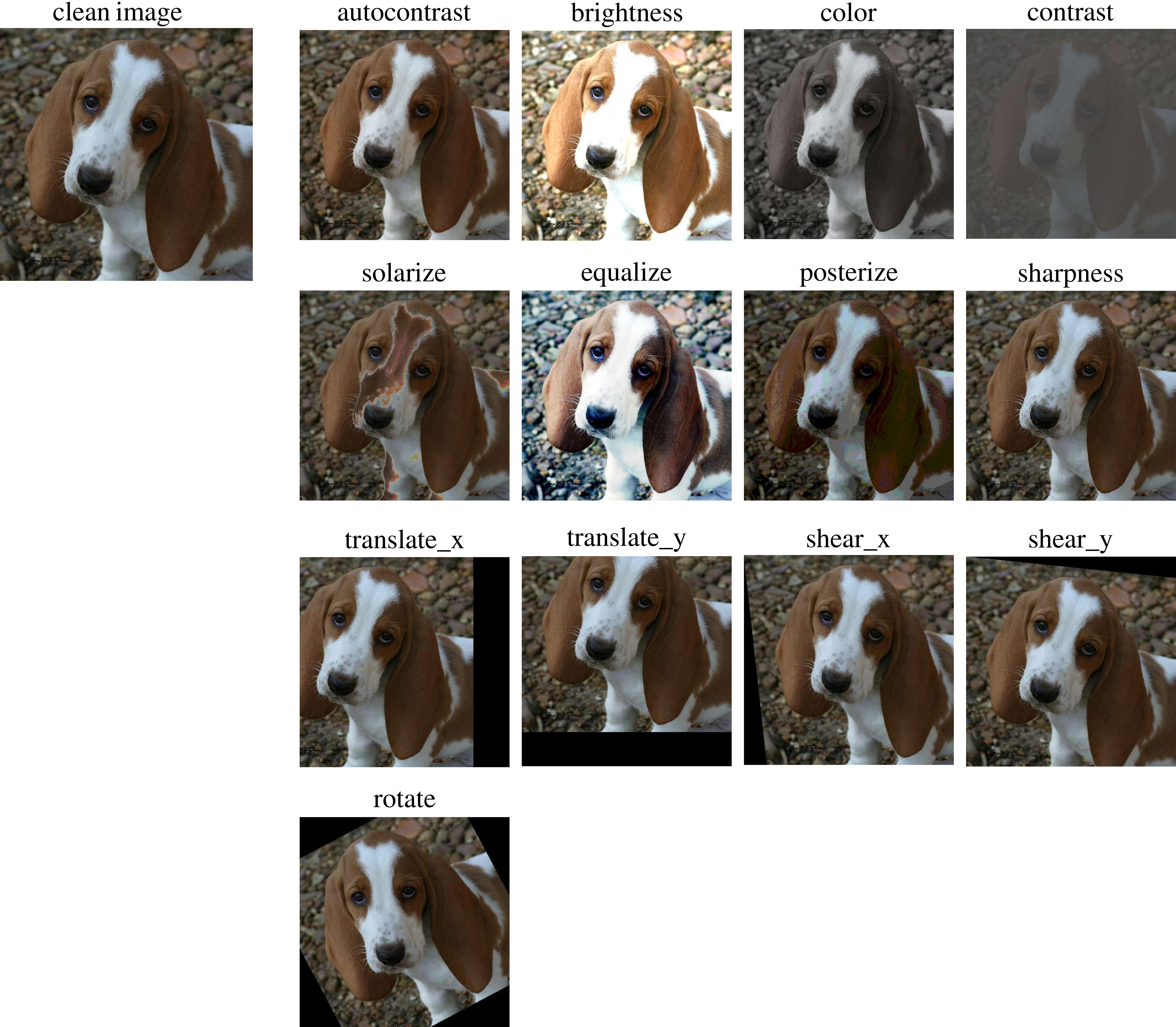}
  \caption{An example of the possible augmentations to be applied on a given input image.}
  \label{fig:all_augs}
\end{figure}

For the set of possible augmentations $\mathcal O$, we follow AutoAugment \cite{cubuk2019autoaugment} and use the augmentations shown in \cref{fig:all_augs}. We note that when mixing augmentations (\ie $K>1$), we also add the possibility of composing augmentations, \eg for $K = 3$, we give the possibility of sampling a pair of augmentations, so that a given of the operation $o_i$ might be composed of two operations  $o_i = o_{i1} \circ o_{i2}$.

For semantic segmentation, we limit $\mathcal O$ to only contain photometric augmentations to avoid changing input coordinate-space, \ie we remove 
\texttt{translate\_x}, \texttt{translate\_y}, \texttt{rotate}, \texttt{shear\_x} and \texttt{shear\_y} from $\mathcal O$. However, it is possible to maintain the geometric transformations and use a bilinear resampler to bring back the outputs of the augmented image into the coordinate-space of the clean image.

\subsubsection{Mean Teacher} The objective of the consistency loss in \cref{eq:target_consistency}.
%\cref{eq:target_consistency}
is to incrementally push the decision boundary to low-density regions on the target domain. However, using the current model $h$ as both a teacher, generating pseudo-labels for the target examples, and as a student, producing the current predictions over perturbed inputs, might result in an unstable training, where a small optimization step can result in a significantly different classifier, hurting the target generalization performance. To solve this, we follow Mean Teachers (MT) \cite{tarvainen2017mean}, and use an Exponential Moving Average (EMA) of the student model $h$ weights as a teacher $h^{\prime}$,
where the weights $\theta^\prime_t$ of the teacher model at a training step $t$ are defined as the EMA of successive student's weights $\theta$:
\begin{equation}
    \theta_{t}^{\prime}=\alpha \theta_{t-1}^{\prime}+(1-\beta) \theta_{t}
\end{equation}
where $\beta$ is a momentum term that controls how far we reache into training history. The teacher model can be used to generate the pseudo-labels $h^{\prime}(\mathbf x^t)$ for a more stable optimization procedure.

\subsubsection{Target Consistency for Semantic Segmentation} To demonstrate the generality of target consistency, we propose to adapt it for segmentation tasks. Given the dense nature of semantic segmentation, where we predict class assignment at each spatial location, we remove the local consistency constraint $\mathcal L_{\mathrm{VAT}}$, since even small perturbations at the pixel level might significantly change the local appearance, making the task of predicting consistent labels impractical. Additionally, we constrain the target consistency to be only photometric augmentations to conserve the input coordinate-space. We follow \cite{tsai2018learning} and adopt adversarial learning in the output space rather than representation space, taking advantage of the structured outputs in semantic segmentation that contain spatial similarities between the source and target domains, the adversarial network is applied at a multi-level to perform output space adaptation at different feature levels effectively. We refer the reader to Section 4 of \cite{tsai2018learning} for more details on multi-level output space-based adaptation.

\subsubsection{Implementation}
For the implementation, we use \texttt{PyTorch} \cite{paszke2019pytorch} deep learning framework and base our implementation on the official implementations of CDAN \cite{long2018conditional}\footnote{\url{https://github.com/thuml/CDAN}} 
and ADVEN \cite{vu2019advent}\footnote{\url{https://github.com/valeoai/ADVENT}}.
All experiments are done on a single NVIDIA V100 GPU with 32GB memory. In terms of the hyperparameters, for classification, we adopt mini-batch SGD with a momentum of 0.9 and the learning rate annealing strategy \cite{ganin2016domain} with an initial learning rate of $10^{-2}$. As for segmentation, the model is trained using mini-batch SGD and a learning rate $2.5 \times10^{-4}$, momentum $0.9$ and weight decay $10^{-4}$, and Adam optimizer \cite{kingma2014adam} for the discriminator with learning rate $10^{-4}$, both with a polynomial learning rate scheduler \cite{chen2017deeplab}.

\subsection{Fourier Analysis of Target Robustness}
\label{sec:fourier}

\begin{figure}[!ht]
    \centering
    \includegraphics[width=0.8\textwidth]{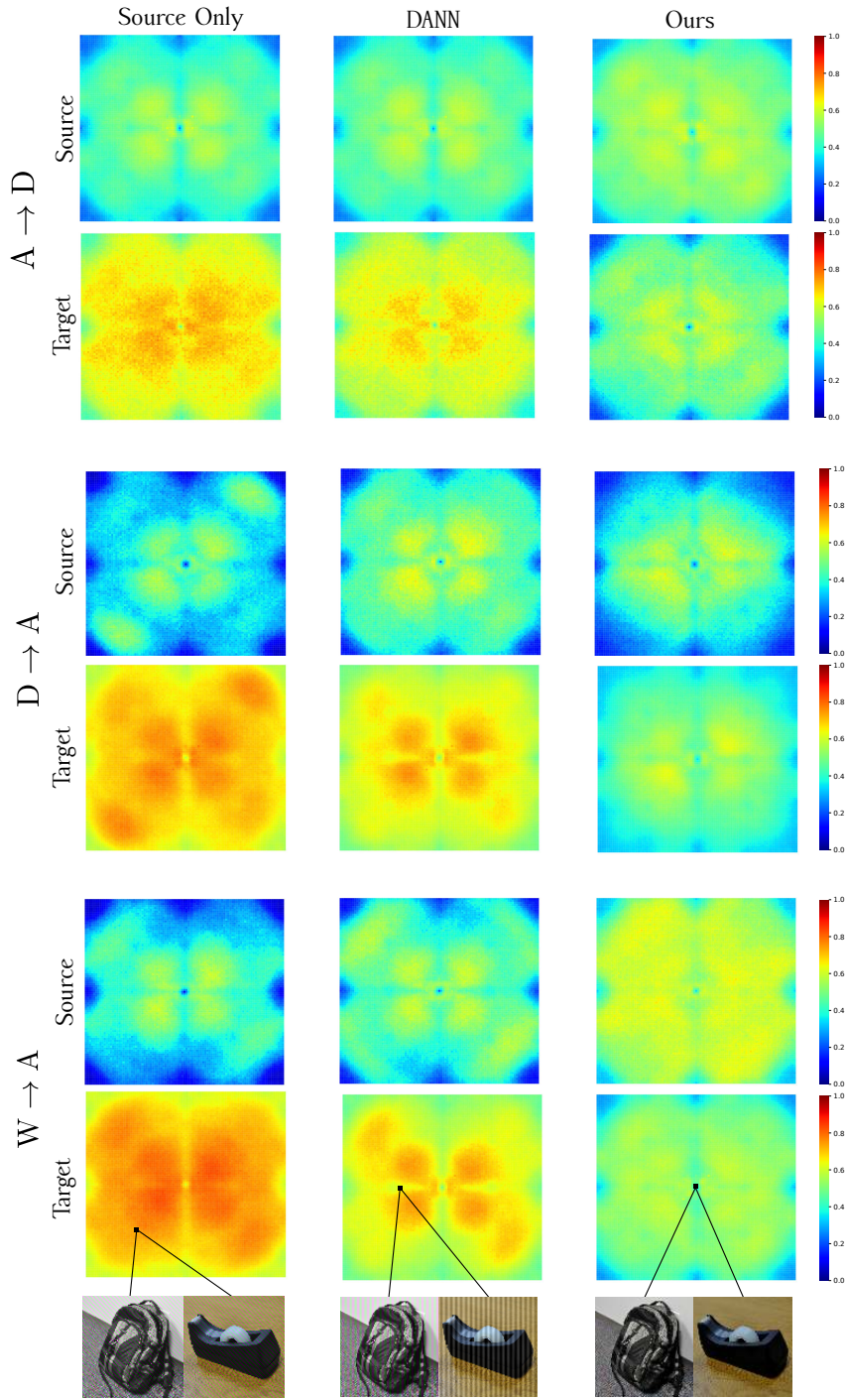}
    \caption{Fourier Analysis of Model Robustness on Source and Target. An illustration of the Fourier sensitivity heatmaps on the source and target domains for a ResNet-50 trained with different objectives. Each pixel of the heatmap is the error of the model when all of its inputs are perturbed with a single Fourier basis vector.}
    \label{fig:fourier}
\end{figure}

To further examine the lack of target robustness in DA, we investigate a common hypothesis in robust deep learning \cite{hendrycks2019augmix}, where the lack of robustness is attributed to spurious high-frequency correlations that exist in the source data, that are not transferable to target data. To his end, we follow \cite{yin2019fourier}, and measure the model error after injecting an additive noise at different frequencies. Concretely, we resize all of the data to $96 \times 96$ images, we then add, at each iteration, $96 \times 96$ Fourier basis vector corresponding to an additive noise at a given frequency, and record the model error over either source or target data when such basis vector is added to each image individually (see Section 2 of \cite{yin2019fourier} for more details). \cref{fig:fourier} shows the Fourier sensitivity heatmaps on source and target, for a ResNet-50 trained with different objectives. Each pixel of $96 \times 96$ heatmaps shows the error of the model when the inputs are perturbed by a single Fourier basis vector, in which the error corresponding to low-frequency noise is shown in the center, and high frequencies are away from the center. We observe that the model is highly robust on source across frequencies and the different objectives, but becomes quite sensitive to high-frequency perturbations on target when trained on source only or with a DANN objective. However, such sensitivity is reduced when enforcing the cluster assumption on the target domain, indicating a possible suppression of the spurious high-frequency correlations found in the source domain.

\subsection{Qualitative Results}
\label{sec:qualitative_res}

\begin{figure}[!ht]
  \centering
  \includegraphics[width=0.9\textwidth]{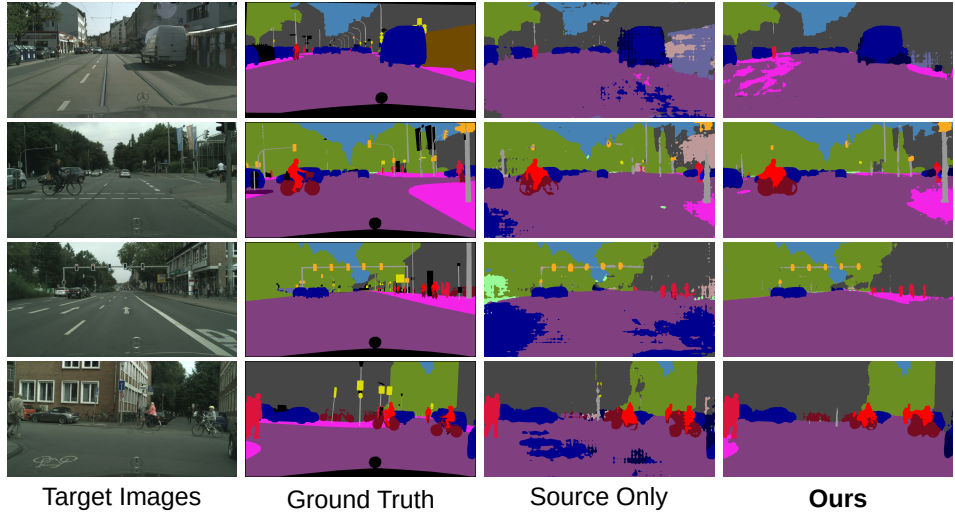}
  \caption{Qualitative Results on GTA5 $\shortrightarrow$ Cityscapes.}
  \label{fig:qulitativeGTA}
\end{figure}

\textbf{Qualitative Results.} \cref{fig:qulitativeGTA} illustrate qualitative results for smantic segmentation.

\begin{figure}[!ht]
    \centering
    \includegraphics[width=0.8\textwidth]{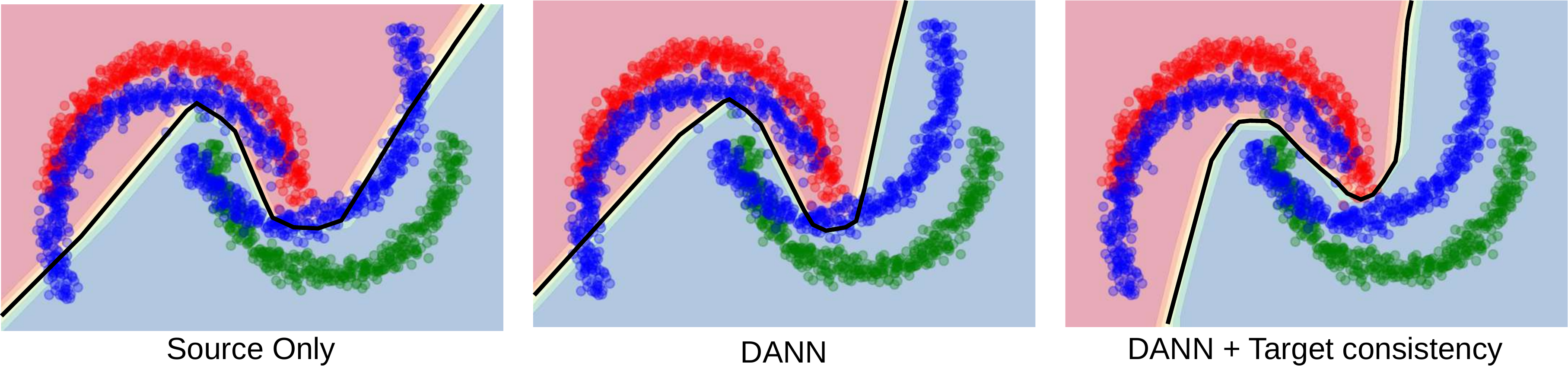}
    \caption{Effect of TC on \textit{two moons} dataset. Red and green points are the instances of the two classes of the source domain. Blue points are target samples generated by rotating source samples. The black line shows the learned decision boundary, when using only source samples, with a DANN objective and with target consistency.}
    \label{fig:twomoons}
\end{figure}

\textbf{Toy Dataset.} To show the effect of TC on the decision boundary, we conduct a toy experiment on the rotating two moons dataset, where the target samples are obtained by rotating the source points by $45^{\circ}$, comparing the learned decision boundary when we train on source only, with a DANN objective, and when using TC. As shown in \cref{fig:twomoons}, the TC terms helps push the decision boundary away from dense target regions, resulting in a well performing prediction function across domains.

%\iffalse
%\begin{figure}[!ht]
%  \centering
%  \includegraphics[width=1\textwidth]{figures/TSNE_all.pdf}
%  \caption{T-SNE of the adapted features of, \textit{left}: ResNet-50, \textit{center}: DANN, \textit{right}: Ours, trained on D $\shortrightarrow$ A task of Office-31. Blue: Source D; Red: Target A.}
%  \label{fig:todo}
%\end{figure}
%\fi

%We visualize the feature representations of \textbf{D $\shortrightarrow$ A}  task of Office-31 with t-SNE \cite{maaten2008visualizing} in \cref{fig:TSNE}. We observe that our method produces a well aligned source and target features. This shows the benefits of coupling consistency regularization with class level discrimination. The proposed method also helps in  obtaining local consistent and globally coherent predictions for semantic segmentation as illustrated in \cref{fig:qualitative_segmentation}.

\subsection{Theory}
\label{sec:theory}
\subsubsection{Non-conservative Domain Adaptation}
\begin{theorem}[From \cite{ben2007analysis} and \cite{ben2010theory}] \label{ben_david} Given a hypothesis class $\mathcal H$ and a hypothesis $h\in\mathcal H$: 
\begin{equation}
    \label{BD_H}
    \varepsilon^t (h) \leq\varepsilon^s(h) + d_{\mathcal H \Delta\mathcal H} + \lambda_{\mathcal H}
\end{equation}
where $d_{\mathcal H \Delta\mathcal H} := \sup_{h,h' \in \mathcal H} | \varepsilon^s(h,h') - \varepsilon^t(h,h')|$ and $\lambda_{\mathcal H} := \inf_{h\in\mathcal H} \{ \varepsilon^t(h) + \varepsilon^s(h)\}$. In particular, provided a representation $\varphi$, and applying the inequality to $\mathcal G\circ\varphi:=\{g \circ \varphi: g \in\mathcal G\}$:
\begin{equation}
    \label{BD_G}
    \varepsilon^t(g\varphi) \leq \varepsilon^s(g\varphi) + d_{\mathcal G \Delta \mathcal G}(\varphi) + \lambda_{\mathcal G}(\varphi)
\end{equation} 
where $d_{\mathcal G \Delta \mathcal G}(\varphi) := \sup_{g,g' \in\mathcal G} |\varepsilon^s(g\varphi) - \varepsilon^t(g'\varphi)|$ and $\lambda_{\mathcal G}(\varphi) := \inf_{g \in \mathcal G} \{ \varepsilon^s(g\varphi) + \varepsilon^t(g\varphi) \}$.
\end{theorem}
On the one hand, \cref{BD_H} shows the role of the hypothesis class capacity for bounding the target risk. The lower the hypothesis class sensitivity to changes in input distribution, the lower $d_{\mathcal H \Delta \mathcal H}$. On the other hand, \cref{BD_G} puts emphasis on representations: if source and target representations are aligned \ie $p(\mathbf z^s) \approx q(\mathbf z^t)$ for $\mathbf z := \varphi(\mathbf x)$, then $d_{\mathcal G \Delta \mathcal G}(\varphi) = 0$.

%Prior works \cite{ganin2015unsupervised, ganin2016domain, long2015learning, long2016unsupervised, long2017deep, long2018conditional} have greatly improved capacity to achieve a trade-off between source classification error and domain invariance of representations by minimizing $\varepsilon^s(g\varphi) + d_{\mathcal G \Delta \mathcal G}(\varphi)$ from \cref{BD_G}. Clearly, maintaining a low $\lambda_{\mathcal G}(\varphi)$ while learning domain invariant representations is a key to success. Some works bring theoretical evidence of its difficulty \cite{zhao2019learning, wu2019domain, johansson2019support} while pioneering works dig into that direction \cite{liu2019transferable, chen2019transferability, wang2019transferable}. This difficulty is referred as \textit{non-conservative} DA in \cite{shu2018dirt} \ie when the optimal joint classifier is significantly different from the target optimal classifier:

One of the main difficulties of DA is achieving the optimal trade-off between source classification error and domain invariance of representations by minimizing $\varepsilon^s(g\varphi) + d_{\mathcal G \Delta \mathcal G}(\varphi)$ from \cref{BD_G}, while maintaining a low $\lambda_{\mathcal G}(\varphi)$. This difficulty is referred as \textit{non-conservative} DA in \cite{shu2018dirt} \ie when the optimal joint classifier is significantly different from the target optimal classifier:

\begin{equation}
\inf_{h \in \mathcal{H}} \varepsilon^t(h)< \varepsilon^t(h^\lambda) \text { where } h^\lambda:=\underset{h \in \mathcal{H}}{\arg \min }\ \varepsilon^s(h)+\varepsilon^t(h)
\label{eq:optimality_gap}
\end{equation}

Non-conservative DA can be described from the point of view of the hypothesis class as described by\cite{shu2018dirt}, \ie \cref{BD_H} from \cref{ben_david}, then allowing change in representations to detect it, \ie $\inf$ computed on $\mathcal H =\mathcal G \circ \Phi$ in \cref{eq:optimality_gap}. Similarly, when provided with a representation $\varphi$, the optimal joint classifier differs from the target optimal classifier: 
\begin{equation}
    \inf_{g\in\mathcal G} \varepsilon^t(g\varphi) < \varepsilon^t(g^\lambda\varphi) \mbox{ where } g^\lambda := \arg \min_{g \in \mathcal G}\{ \varepsilon^s(g\varphi) + \varepsilon^t(g \varphi) \}
\end{equation}
This expression reflects the view of the literature of domain adversarial learning which puts emphasis on representations, \ie \cref{BD_G} from theorem \cref{ben_david}. Note this definition only allows to modify the classifier, $\inf$ computed on $\mathcal G$, for detecting non-conservative DA, which may be a weak indication. We extend the denomination of non conservative DA to the case where $\varepsilon^t(\varphi):=\inf_{g\in\mathcal G} \varepsilon^t(g\varphi)$ is not optimal in $\varphi$.

\subsection{Theoretical Analysis}

We provide theoretical insights into the interaction between TC and class-level invariance. We consider $\varphi \in \Phi$ and $g \in \mathcal G$, which are modified to obtain $\tilde \varphi$ and $\tilde g$ defined as the closest instances such that $\tilde g\tilde \varphi$ verifies TC. For instance, they can be obtained by minimizing  $ \ell_2(\varphi, \tilde \varphi)  +  \ell_2( g , \tilde g)  +  \lambda \cdot \mathcal L_{\mathrm{TC}}(\tilde g, \tilde \varphi)$ where $\ell_2$ is an $L^2$ error. When enforcing TC, we expect to decrease the target error \ie $
    \varepsilon^t (\tilde g \tilde \varphi) < \varepsilon^t(g\varphi)  $. Noting $\rho := (1- \varepsilon^t (\tilde g \tilde \varphi) / \varepsilon^t(g\varphi))^{-1}$ and $\tilde{\mathbf y} := \tilde g \tilde\varphi(\mathbf x)$, $\mathsf F$ a large enough critic function space, we adapt the theoretical analysis from \cite{bouvier2020robust}: 
\begin{equation}
    \label{bound_final}
    \varepsilon^t( g \varphi) \leq \rho \left ( \varepsilon^s(g\varphi) + 8\sup_{\mathsf f \in \mathcal F} \left \{ \mathbb E_{(\mathbf z^s, \mathbf y^s) \sim p}[ \mathbf y^s \cdot \mathsf f(\mathbf z^s)] - \mathbb E_{(\mathbf z, \tilde{\mathbf y}) \sim q}[ \tilde{ \mathbf y}^t \cdot \mathsf f(\mathbf z^t)] \right\}  + \inf_{\mathsf f \in \mathcal F} \varepsilon^t(\mathsf f \varphi) \right )
\end{equation}
More precisely, $\mathsf F$ has the following properties \cite{bouvier2020robust}: 

\begin{itemize}
    \item (A1) $\mathsf F$ is symmetric (\textit{i.e.} $\forall f \in \mathsf F, -f \in \mathsf F$) and convex.
    \item (A2) $\mathcal G \subset \mathsf F$ and $\left  \{\mathsf f \cdot \mathsf f'~ ; ~\mathsf f, \mathsf f' \in \mathsf F \right \} \subset \mathsf F$.
    \item (A3) $\forall \varphi \in \Phi$, $\mathsf f_D(z) \mapsto \mathbb E_D[Y|\varphi(X)=z] \in \mathsf F$. 
    \item (A4) For two distributions $p$ and $q$ on $\mathcal Z$, $p=q$, and $1 \leq c \leq C$, if and only if,
    \begin{equation}
        \mathrm{IPM}(p,q ; \mathsf F_c) := \sup_{\mathsf f \in \mathsf F} \left \{ \mathbb E_p[\mathsf f_c(Z)] - \mathbb E_q[\mathsf f_c(Z)] \right \} = 0
    \end{equation}
    where $\mathsf f_c$ is the $c-$th coordinate of $\mathsf f$.
\end{itemize}

Crucially, by observing that $\sup_{\mathsf f \in \mathcal F} \left \{ \mathbb E_{(\mathbf z^s, \mathbf y^s) \sim p}[ \mathbf y^s \cdot \mathsf f(\mathbf z^s)] - \mathbb E_{(\mathbf z, \tilde{\mathbf y}) \sim q}[ \tilde{ \mathbf y}^t \cdot \mathsf f(\mathbf z^t)] \right\}$ is an \textit{Integral Probability Measure} proxy of $\mathcal L_{\mathrm{CLIV}}$, \cref{bound_final} reveals that class-level domain invariant representations can leverage feedback from an additional regularization, here the Target Consistency, to learn more transferable invariant representations. 

\cref{bound_final} is obtained by applying Bound 4 (Inductive Bias and Guarantee, equation 14) from \cite{bouvier2020robust} by leveraging the inductive design: 
\begin{equation}
    \varepsilon^t (\tilde g \tilde \varphi) < \varepsilon^t(g\varphi)
\end{equation}
provided by the Target Consistency. Note that, following the notations of \cite{bouvier2020robust}, we have bounded the INV term (defined in equation 8 \cite{bouvier2020robust}) by the TSF term (defined in equation 9 \cite{bouvier2020robust}) leading to 6TSF, were $\mathrm{TSF}:= \sup_{\mathsf f \in \mathcal F} \left \{ \mathbb E_{(\mathbf z^s, \mathbf y^s) \sim p}[ \mathbf y^s \cdot \mathsf f(\mathbf z^s)] - \mathbb E_{(\mathbf z, \tilde{\mathbf y}) \sim q}[ \tilde{ \mathbf y}^t \cdot \mathsf f(\mathbf z^t)] \right\}$ in our case. Besides, the constant term is $\rho := 1/(1- \beta)$, not $\beta / (1-\beta)$, since we bound $\varepsilon^t(g\varphi)$ not $\varepsilon^t(\tilde g \tilde \varphi)$ where $\beta := \varepsilon^t(\tilde g \tilde \varphi) / \varepsilon^t(g\varphi)$.

\end{document}